\documentclass{ws-ijcis}

\usepackage[T1]{fontenc}
\usepackage{graphicx}
\usepackage{xcolor, soul}
\usepackage{colortbl}
\usepackage{hyperref}
\usepackage{color}
\usepackage{algorithm}
\usepackage{algpseudocode}
\usepackage{multirow}
\usepackage{wrapfig}
\usepackage{todonotes}
\usepackage{amsmath}
\usepackage{amssymb}
\usepackage{array}
\usepackage{enumitem}
\usepackage{longtable}
\usepackage{tabularx}
\usepackage{makecell}
\usepackage{longtable}
\usepackage{xltabular}
\usepackage{xtab}
\usepackage[most]{tcolorbox}
\usepackage{pifont}
\usepackage{circledsteps}
\usepackage{xcolor}
\definecolor{ForestGreen}{RGB}{34,139,34}

\newcommand{\rom}[1]{%
  \textup{\uppercase\expandafter{\romannumeral#1}}%
}

\definecolor{mylila1}{RGB}{232, 179, 233}
\definecolor{mylila2}{RGB}{192, 97, 203}
\definecolor{myblue2}{RGB}{98, 160, 234}

\newcommand\nk[1]{\textcolor{black}{#1}}
\newcommand\rnk[1]{\textcolor{black}{#1}}
\newcounter{tfnumquote}

\newcounter{tnumquote}

\newcommand\nkMulti{\color{black}}

\renewcommand{\footnotesize}{\scriptsize} 

\newcommand\srm[1]{\textcolor{black}{#1}}
\newcommand\srmrev[1]{\textcolor{black}{#1}}

\newcommand\newnk[1]{\textcolor{black}{#1}}
\begin{document}

\markboth{Klievtsova et al.}{Conversational Process Model Redesign}
\catchline{0}{0}{0000}{}{}

\title{Conversational Process Model Redesign}

\author{Nataliia Klievtsova}
\address{Technical University of Munich,\\TUM School of Computation, Information and Technology,\\Boltzmannstraße 3
85748 Garching b. München, Germany}

\author{Timotheus Kampik}
\address{SAP Signavio,\\George-Stephenson-Straße 7-13, 10557 Berlin, Germany}

\author{Juergen Mangler}
\address{Technical University of Munich,\\TUM School of Computation, Information and Technology,\\Boltzmannstraße 3
85748 Garching b. München, Germany}

\author{Stefanie Rinderle-Ma}
\address{Technical University of Munich,\\TUM School of Computation, Information and Technology,\\Boltzmannstraße 3
85748 Garching b. München, Germany}

\maketitle              

\begin{abstract}
With the recent success of large language models (LLMs), the idea of AI-augmented Business Process Management systems is becoming more feasible. One of their essential characteristics is the ability to be conversationally actionable, allowing humans to interact \srm{with the LLM effectively to perform crucial process life cycle tasks such as process model design and redesign.} However, most current research focuses on single-prompt execution and evaluation of results, rather than on continuous interaction between the user and the LLM. In this work, we aim to explore the feasibility of using LLMs to empower domain experts in the creation and redesign of process models in an iterative and effective way. The proposed conversational process model redesign (CPMR) approach receives as input a process model and a redesign request by the user in natural language. Instead of just letting the LLM make changes, the LLM is employed to (a) identify process change patterns from literature, (b) re-phrase the change request to be aligned with an expected wording for the identified pattern (i.e., the meaning), and then to (c) apply the meaning of the change to the process model. This multi-step approach allows for explainable and reproducible changes. In order to ensure the feasibility of the CPMR approach, and to find out how well the patterns from literature can be handled by the LLM, we perform an extensive evaluation\srmrev{, also in comparison to a baseline approach without change patterns.} The results show that some patterns are hard to understand by LLMs and by users \srmrev{ and that clear change descriptions by users are essential. Overall, we recommend a hybrid approach that identifies all used change patterns and then directly applies those patterns that work correctly and for the others derives follow-up questions in order to improve user input.}

\end{abstract}

\section{Introduction}
\label{sec:intro}

Business process modelling is an approach to describe how businesses execute their operations~\cite{not} by using graphical constructs to specify the business logic.
The utilization of a standardized notation such as Business Process Model and Notation (BPMN 2.0\footnote{\url{www.omg.org/spec/BPMN/2.0}}) typically improves operational efficiency, significantly minimizes errors, and enhances communication and collaboration. One of the primary challenges is the extensive training and skill development required for best-practice utilization of BPMN by various stakeholders within an organization, such as domain experts and process designers/modellers. The successful creation of best-practice models~\cite{role} can be facilitated either by extensive collaboration between domain experts and modellers, or by investing in training programs for domain experts, so that they can handle modelling tasks themselves.

While collaborations help to avoid the implementation of special training programs and ensure that BPMN models are well designed~\cite{role}, they can also lead to a ``dilemma between process modeller and domain expert'' as there is no or only limited knowledge overlap between them, i.e., there exists a communication gap. The process modeller lacks specific domain knowledge, while the domain expert may have only limited knowledge of process modelling notations~\cite{Approximating}.
The constant need to transfer the domain knowledge to process modellers is especially burdensome for organizations continuously undergoing adaptations caused by internal or external changes, i.e., when business processes need to be designed or redesigned to improve their day-to-day execution performance~\cite{Beverungen14}. Hence, it is crucial to find a simple and effective way to generate, manipulate, and evaluate process models, minimizing the communication effort of domain experts.

\begin{figure}[htb!]
    \centering
    \includegraphics[width=\linewidth]{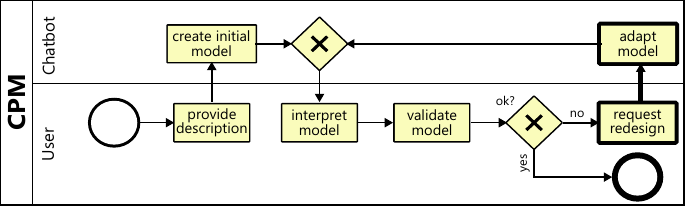}
    \caption{Conversational process modeling including conversational process model redesign}
    \label{fig:cpm}
\end{figure}

\textsl{Conversational process modelling (CPM)}~\cite{convermod,klievtsovawi2024} aims to maximize the involvement of domain experts in the creation of process models and hence to minimize the communication effort between domain experts and process modellers~\cite{LeopoldMP14}. Specifically, CPM refers to the iterative process as depicted in Fig. \ref{fig:cpm} of creating process models based on process descriptions and conversations between users and chatbots, until the created models reach a certain quality level and become sufficiently mature to fulfil their purpose. \srm{This possibly includes several \textsl{process redesign} cycles in which the current process model is changed according to user redesign requests (see tasks with bold lines in Fig. \ref{fig:cpm}) by the LLM. We refer to this as \textsl{conversational process model redesign (CPMR)}. }

\srm{To the best of our knowledge, no CPMR approach has been proposed so far. Hence, this work aims at i) providing a CPMR approach and ii) evaluating the quality of the redesigned models created through this approach. The basic idea of the presented CPMR approach is to pass a user redesign request in natural language (text) to the LLM and to equip the corresponding prompts with well-established change patterns from the literature \cite{DBLP:journals/dke/WeberRR08}. The assumption is that the redesign requests stated by the users involve adaptions of the process model, e.g., inserting a new task or deleting an existing one, that can be represented based on change patterns. The first study presented in \cite{coopis} showed that redesign requests actually involve change patterns. However, the initial assumption of \cite{coopis}, i.e., that the requests mainly refer to basic change patterns such as insert or delete, did not hold. In fact, redesign requests might also involve more complex change patterns such as embedding tasks into loops. \srmrev{Moreover, the performance of CPMR has to be evaluated in comparison with the baseline of LLM-based process model redesign without involving any change patterns. } Hence, this work extends the previous work presented in \cite{coopis} with a systematic and comprehensive evaluation of (i) all $14$ change patterns presented in \cite{weber}, \nk{(ii) additional patterns that are not part of the original set but emerged as relevant extension to support more complex or diverse process redesign needs}, \srmrev{and a comparison with the baseline approach without change patterns}. To this end, the change patterns are analysed for usage and representation in a conversational context. Moreover, the CPMR approach is formalized and evaluation concepts are provided. The evaluation concepts include the creation of several redesigned process models, i.e., one created by the LLM based on change patterns \nk{and user input}, and one created manually using change patterns for comparison reasons.} \srmrev{Overall, the results recommend a \textsl{hybrid} CPMR approach by first identifying all used change patterns, followed by the direct application of those patterns that work well. For the other patterns, follow-up questions are derived in order to improve user input.}

The paper is structured as follows:
Section~\ref{sec:redesign} describes the CPMR approach and the evaluation concepts. Section \ref{sec:eval} puts existing change patterns into a conversational context. Section~\ref{sec:evaluation} presents the results of a user study to assess the quality of the LLM-redesigned models regarding user satisfaction, model completeness and correctness, layouting, and the quality of the selected graphical representation.  Section~\ref{sec:rewo} discusses related work before
Section \ref{sec:conclusion} concludes the paper.

\section{Conversational Process Model Redesign}
\label{sec:redesign}

This section provides the conceptual framework for conversational process model redesign (CPMR). Section \ref{sub:overview} provides an overview of the approach, \srmrev{including the prompts and Sect. \ref{sec:patterns} the conversational representation of the change patterns. }

\subsection{Overview}
\label{sub:overview}

\begin{figure}[htb!]
    \centering
    \includegraphics[width=1\textwidth]{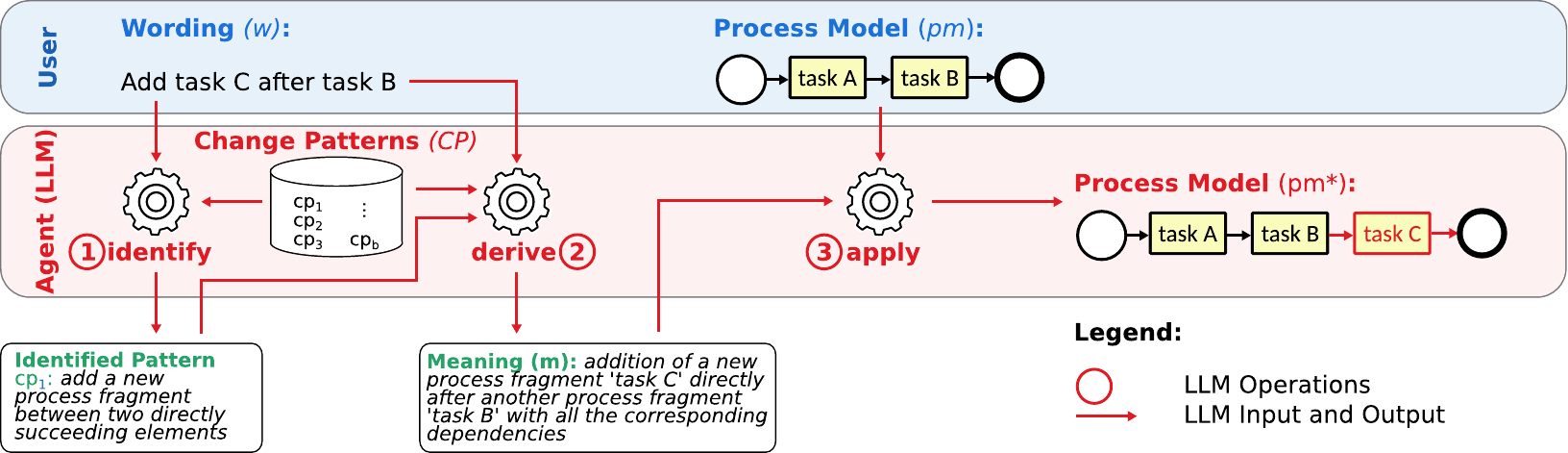}
    \caption{Overview on LLM-based Conversational Process Model Redesign}
    \label{fig:overview}
\end{figure}

The CPMR problem can be formulated as follows: \textsl{Given a process model, a user specifies a redesign request in natural language and the model is adapted based on the request by an LLM.}
The conversational redesign approach tackling this problem is depicted in Fig. \ref{fig:overview}. Its basic idea is to structure the processing of the redesign request by the LLM, exploiting existing and well-established process change patterns as described in \cite{DBLP:journals/dke/WeberRR08}.

\rnk{Based on the results obtained from~\cite{coopis}, the approach is implemented to realize only one change at a time (i.e., one change pattern). We also expect users to address tasks by their labels. Therefore, our approach handles process models under the assumption that all labels are unique.}

\rnk{Change patterns describe recurring modifications or transformations that occur within a business process model.}
\srm{They have a formal semantics as defined in \cite{DBLP:conf/er/Rinderle-MaRW08a}. Assume that a process model $pm$ is transformed into process model $pm^{*}$ by applying change pattern $cp$. Then the formal semantics of $pm$ guarantees that if $pm$ was sound regarding structure and behaviour\footnote{Soundness of process models typically requires certain structural properties such as connectedness and behavioural soundness requires reachability of an end state, etc. For details see \cite{DBLP:books/sp/Weske19}.}, then $pm^{*}$ is sound. }

\srm{Weber et al. \cite{weber} present $14$ patterns, i.e., Insert Process Fragment (cp$_1$), Delete Process Fragment (cp$_2$), Move Process Fragment (cp$_3$), Replace Process Fragment (cp$_4$), Swap Process Fragments (cp$_5$), Extract Subprocess (cp$_6$), Inline Subprocess (cp$_7$), Embed Process Fragment in Loop (cp$_8$), Parallelize Process Fragments (cp$_9$), Embed Process Fragment in Conditional Branch (cp$_{10}$), Add Control Dependency (cp$_{11}$), Remove Control Dependency (cp$_{12}$), Update Condition (cp$_{13}$), and Copy Process Fragment (cp$_{14}$). }

Since most of the patterns can be realised through the \textit{Insert} and \textit{Delete} patterns, the question arises \srm{which of the $14$ change patterns are relevant in the context of CPMR and how they are supported in interactions between user and LLM.}

\srm{We conducted a preliminary survey with $10$ users in~\cite{coopis} which tested the support of change patterns cp$_1$, cp$_2$, cp$_9$, and cp$_{10}$. We opted for these patterns as we assumed them as fundamental for process redesign. } We found that none of the $10$ users referred to the \textit{Insert} and \textit{Delete} patterns (cp$_1$ and cp$_2$), but frequently referred to patterns cp$_8$ (\textit{Add Loop}), cp$_9$, and cp$_{13}$ \textit{(Change Condition)}. Furthermore, users referred to splitting of existing activities into several new activities, which could either be considered a special case of the \textit{Replace} pattern (cp$_4$) or treated as a stand-alone \textit{Split Process Fragment} pattern.

\srm{Based on these results and due to }the fact that LLMs struggle to identify more complex constructs related to the elements and relationships between these elements, we will systematically assess the $14$ change patterns regarding
\begin{enumerate}
    \item[(a)] which redesign requests in textual form (part of the prompt) can be matched to change patterns and if they cannot be matched find possible interpretations.
    \item[(b)] different redesign requests (i.e., do they convey the same meaning, and thus refer to the same change pattern).
    \item[(c)]  operational stability, i.e., if a meaning derived from the redesign request is applied by an LLM to a certain input model, is the output always structurally and semantically identical?
\end{enumerate}

\srm{For tackling (a), we analyse how the change patterns as presented in \cite{weber} can be utilised in a conversational setting, i.e., with textual input instead of the original ``drag \& drop'' manner. The results of this analysis are presented in Sect. \ref{sec:patterns}. In order to describe the LLM-based tasks in semiformal way, assume the following definitions. Let $\cal{W}$ be the set of all wordings where a wording describes a redesign request by a user in natural language. Let PM be the set of all process models. Note that in the following, we assume an imperatively modelled process model in Business Process Modelling and Notation (BPMN) as the de-facto standard. We plan the investigation of declarative process models in future work. Let further CP be the set of all change patterns as described in \cite{DBLP:journals/dke/WeberRR08}. Finally, let $\cal{M}$ be the set of meanings, i.e., parametrised change patterns. Subsequently, we will describe the change patterns in more detail.}

\srm{The user provides a redesign request using \textsl{wording} $w$, e.g., \texttt{Add task C after task B}. Wording w is then passed to the LLM which checks whether a corresponding change pattern in the set of change patterns CP exists. This is realised by function {\Circled[inner color=red, outer color=red]{1} identify}, formally:}
\begin{equation}
 identify: {\cal W} \times 2^{CP} \mapsto CP \cup \{\text{false}\}
\end{equation}

\srm{If the LLM identifies a change pattern $cp$ from $CP$, it returns $cp$. In the example, $cp_1$ for inserting new process fragments between directly succeeding process elements. If no change pattern can be identified, identify returns false. }

\srm{If a predicted change pattern can be identified, i.e., $identify(w,CP)\neq false$, the LLM {\Circled[inner color=red, outer color=red]{2} derives} a \textsl{meaning} $m$ based on wording $w$ and change pattern $cp$, formally:
\begin{equation}
derive: CP \times \cal{W} \mapsto \cal{M} \cup \{\text{false}\}
\end{equation}
Meaning $m$ can be understood as change pattern $cp$ parametrised with the redesign request - wording $w$.}

\srm{The LLM \Circled[inner color=red, outer color=red]{3} {\color{red} applies} the parametrised change pattern $m$ to the \textsl{process model} $pm$ to be redesigned, i.e., $pm$ is transformed into $pm^*$ which can be assumed to be sound, formally:}

\begin{equation}
    apply: \cal{M} \times PM \mapsto PM
\end{equation}

\textbf{Prompt Engineering.}
To enable the agent to perform the functions \texttt{identify}, \texttt{derive}, and \texttt{apply}, we rely on prompt engineering (i.e., instructions that guide the behaviour of the LLM), where each function is implemented as a distinct prompt template. Similar to the baseline approach, each prompt consists of two parts: system instructions on one side and user input on the other.

\textbf{Identify.} This prompt is necessary for classifying the redesign request into one of the predefined change patterns for business process model redesign. The list of available change patterns, which is part of the system input, and the wording provided by the user are components of the user input.

\begin{tcolorbox}[title=Identify,
title filled=false,
colback=mylila2!5!white,
colframe=mylila2!75!black,
breakable]

\textbf{System Prompt:} You are an expert in BPMN (Business Process Model and Notation) modeling. Your task is to evaluate and interpret user-provided modifications to a BPMN process model.

Your task is to classify the user input into one of the predefined change patterns for process model redesign, if a matching pattern exists.
Use the following classification of change patterns to interpret user modifications: 
\texttt{<List of Existing Change Patterns>}.

If a match is found, return only the pattern ID. Only one pattern can be matched.
If no match is found, return NA. No other information is allowed to be returned!!!

\textbf{User Prompt:} \texttt{<Wording provided by a user>}.
\end{tcolorbox}

If the pattern is identified (i.e., the wording corresponds to one and only one existing change pattern), the agent returns only the pattern ID (e.g., ``cp1''). If no match is found (e.g., multiple patterns match, the wording is unclear, misleading, etc.), return ``NA''.

\textbf{Derive.} The main task of this prompt is to derive the meaning by parametrizing the change pattern with the redesign request and ensuring that it fits the BPMN modelling rules. 
\rnk{Based on the identified change pattern from the previous step, the identified pattern ID and its description extracted from the list of available change patterns is incorporated into the user input along with the wording provided by the user. This results in a new prompt.}

\begin{tcolorbox}[title=Derive,
title filled=false,
colback=mylila2!5!white,
colframe=mylila2!75!black,
breakable]

{\nkMulti
\textbf{System Prompt:} You are an expert in BPMN (Business Process Model and Notation) modeling. Your task is to evaluate and interpret modifications to a BPMN process model. The user will provide an input modification based on a predefined change pattern.
Your responsibilities are:

    \textbf{(a)} Validate whether the user-provided input modification contains enough unambiguous information to apply the predefined change pattern.

    \textbf{(b)} Interpret the meaning of the modification based on BPMN semantics and the predefined change pattern.

    \textbf{(c)} Ensure the modification complies with BPMN modeling rules and fits within the structure of the existing process.

Return only the clear meaning of the modification in natural language, without any ambiguity or additional information. If the input does not contain sufficient details to apply the change pattern, return "NA".

\textbf{User Prompt:} Identified changed pattern is \texttt{<Pattern ID>} -  \texttt{<Pattern Description>}. Changes applied to the model: \texttt{<Wording provided by a user>}.
}
\end{tcolorbox}

If it is possible to derive the meaning, the agent returns the meaning in the form of text expressed in natural language. If, for some reasons, it is not possible to derive the meaning, the agent returns ``NA''.

\textbf{Apply.} The aim of this prompt is to apply the changes to the input process model pm based on the provided meaning m, while adhering to the syntax rules of the desired output format. The syntax rules of the desired output format are the part of the system input, and the input process model along with the meaning are components of the user input. 

\begin{tcolorbox}[title=Apply,
title filled=false,
colback=mylila2!5!white,
colframe=mylila2!75!black, 
breakable]

\textbf{System Prompt:} You are an expert in BPMN modelling, specifically in \texttt{<Output Format>} format.
Your task is to validate and transform BPMN models based on user-provided modifications, ensuring compliance with BPMN rules and \texttt{<Output Format>} syntax.
You are allowed to adjust only those parts of the process model mentioned in the user-provided modification. Other parts of the model have to stay unchanged.

The \texttt{<Output Format>} syntax for BPMN models is described as follows:
\texttt{<Rules for the Process Model in Output Format >}.

Return only \texttt{<Output Format>} as text without any additional information! Give me just the raw \texttt{<Output Format>} code without markdown formatting.

\textbf{User Prompt:} Consider following process model: \texttt{<Input Process Model>}.
Apply these changes to the model: \texttt{<Meaning>}.
\end{tcolorbox}

{\nkMulti
To create the \textsl{Apply} prompt, we utilize a \textit{zero-shot} prompting strategy (providing solely task instructions without any specific examples), even though the accuracy of zero-shot prompting is generally lower than the accuracy of few-shot prompting~\cite{debnath2025comprehensive}. We use zero-shot prompting as it relies on LLM knowledge gained during the training phase, demonstrating the LLM's understanding of a particular task ``AS-IS''.

\newnk{Zero-shot prompting and its performance can provide an informative baseline, demonstrating strengths and weaknesses of LLMs inherited solely from pre-trained knowledge. This is particularly relevant in our case, where multiple patterns exist and a single pattern can be expressed in various ways. Zero-shot prompting allows to evaluate current capabilities of LLMs, estimate their understanding of the situation, compare multiple LLMs with each other, and identify points for further improvement (e.g., \emph{Is the information we currently provide sufficient?} \emph{Should we improve or elaborate on the description of all patterns or only specific ones?} \emph{Do we need to provide examples for each pattern or just one?} \emph{Is the current task granularity adequate, or should we explore other options?} \emph{Should we switch to a more advanced prompting strategy such as chain-of-thought?}). Insights from zero-shot prompting can be incorporated into few-shot prompting to emphasise correct examples, improve their design, and optimize the overall quality of the generated output.}

In our case, we rely on the same prompting strategy used to generate process models with LLMs from natural language text. As we know that full model creation is possible with the zero-shot strategy, we can use this technique to estimate (1) whether the LLM can make small changes reliably, (2) which changes can or cannot be performed easily, and (3) whether all LLMs demonstrate the same performance tendencies across multiple process model changes. 

The number of examples given to the LLM can influence its performance in two ways: (a) a recency bias can occur (the LLM repeats the answer in the last example instead of really solving the task)~\cite{guo2024makes} or (b) examples are interpreted as a part of instruction~\cite{examples} leading to misinterpretation of actual desired change. 

Another factor to consider is that by providing multiple examples the length of the input provided to the LLM is increased. The performance of current LLMs drops substantially with both longer inputs and outputs, even before reaching the maximum input or output length capacity of the LLMs~\cite{input,output} due to the inability to identify relevant information in the long inputs~\cite{du2025contextlengthhurtsllm}.
}

\subsection{ Conversational Representation of Change Patterns}
\label{sec:patterns}

\rnk{Change patterns serve as a foundation of the proposed CPMR approach.} \srm{Thus, we elaborate on how to utilise change patterns in a conversational setting. The reason is that the change patterns presented in \cite{weber} were designed with the assumption (or expectation) that the user interacts with the modelling environment through a graphical user interface. In our case, users interact with the LLM in the CPMR approach using as text rather than relying on the typical drag-and-drop behaviour.
Communicating with the LLM agent through a conversational user interface (CUI) might provide a more natural and engaging user experience, and as a result, the users are less restricted in their functionality~\cite{guivscui}. This can be seen as an advantage or a disadvantage, as redesign requests can express the same patterns in multiple ways or even exceed the system's intended scope~\cite{10.1145/3447526.3472036}.
Consequently, it is possible that the existing low-level primitives and high-level patterns are insufficient or not required for conversational interaction with the LLM and for navigating the process model.}

Therefore, in this section, we introduce and summarise additional patterns, that can serve as extensions to the existing change patterns proposed in~\cite{weber}, based on both existing literature (see Sect.~\ref{sec:rewo}) and personal experience gained through interaction with an LLM-agent via the CUI.

\nk{As mentioned in Section~\ref{sec:rewo}, change patterns are a combination of simple actions on individual model elements, leading to process model modifications. These simple actions (i.e., often called \textit{low-level primitives}) each refer only to a single process model element at a time and have no structural assumptions about the model~\cite{weber}. Typical low-level primitives are: add node, delete node, add edge, remove edge, and move edge~\cite{xydiscomparing}. The problem with these primitives is that when performing more complex changes like, for instance, adding a new task into a process, we are required to add not only a node (i.e., task), but also all the corresponding edges that connect this new element with the existing process model~\cite{DBLP:journals/dke/WeberRR08}.}

\nk{Since a simple operation like ``add task'' consists of at least three primitives, like adding a node and adding two edges, operating with low-level primitives quickly accelerates in complexity when applying more complicated changes, and can lead to multiple potential errors even for users who have sufficient modelling skills. Thus, to support users in model redesign, more complex high-level patterns are used. Unlike primitives, high-level change patterns require an understanding of the model’s structure and its modelling rules.}

Considering existing \textbf{low-level primitives} and assuming that the model is generated by the agent, it may be necessary to change the label of an element so that it better aligns with user needs. Therefore, we state that it is necessary to introduce an additional low-level primitive—\textbf{``rename node''}. Renaming a node is considered a simple low-level primitive because it does not manipulate the structure or flow of the process model but directly modifies an individual element at the semantic level. Additionally, this primitive cannot be replaced by any other. While one might argue that renaming a node could be achieved by combining the ``remove node'' and ``add node'' primitives, this approach results in the substitution of one element with another. In contrast, applying ``rename node'' ensures that the element remains the same, with no changes to any of its other properties (if such exist), except for its label.

Analysing existing \textbf{high-level patterns} we first exclude patterns that are not supported within the BPMN 2.0 standard \newnk{since originally the patterns were defined on a more abstract, universal process meta-model that does not always correspond to BPMN constructs}: patterns cp$_{11}$ and cp$_{12}$ (Add and Delete Control Dependencies). Secondly, we propose splitting pattern \textbf{cp$_8$} (Embed Process Fragment in Loop) into two separate patterns: \textbf{cp$_{8.1}$} for Pre-conditional Loop and \textbf{cp$_{8.2}$} for Post-conditional Loop since their behaviours have to be expressed in natural language in a different way. Although both patterns express the same intent -- to execute a particular process fragment multiple times based on a condition -- the timing of the condition check results in different behaviour within the model. In \textbf{cp$_{8.2}$}, a task must be executed at least once, whereas in \textbf{cp$_{8.1}$}, a task can either be executed multiple times (similar to cp$_{8.2}$) or not at all. This variation in behaviour within the same pattern leads to different instructions in wording and meaning and, consequently, differences in output Model. Therefore, the division into two distinct patterns is required.

In addition, we define five patterns that are considered to be required during CPMR: Split Process Fragment (cp$_{15}$), Merge Process Fragment (cp$_{16}$), Delete Entire Branch (cp$_{17}$), Leave Single Branch (cp$_{18}$), and Replace Gateways (cp$_{19}$).

\textbf{cp$_{15}$} (Split Process Fragment) allows splitting an existing process fragment into multiple separate process fragments. It is an efficient way to separate tasks, as it also adjusts the control flow between the split tasks. This pattern is useful when multiple tasks that should be performed sequentially are currently combined into a single task but these tasks do not have enough complexity or structure to justify forming a separate subprocess.

This pattern affects the granularity of the process model. It differs from cp$_7$ (Inline Subprocess), which expands an already structured subprocess, making its tasks visible in the main process. In contrast, cp$_{15}$ increases granularity but does not necessarily create a subprocess.

For example, receiving a document and reviewing it for approval might be combined into a single task, ``Receive and Review Document''. However, since these two activities are sequential and need to be tracked separately, cp$_{15}$ would split them into two distinct tasks: ``Receive Document'' and ``Review Document''.

\textbf{cp$_{16}$} (Merge Process Fragment) serves for merging multiple existing separate process fragments into one process fragment. It is efficient for merging tasks in a single task, as it also adjusts the control flow. This pattern is useful when activities that are represented separately and considered to be independent are actually a single activity or is a single activity from stakeholder perspective.

This pattern is fundamentally different from cp$_6$, as cp$_{16}$ combines multiple independent process fragments into a single process fragment within the process. cp$_6$, on the other hand, takes a set of related process fragments and moves them into a separate subprocess, creating a structured, reusable component.

For example, ``Generate Invoice'', ``Verify Invoice Details'' and ``Send Invoice'' from the perspective of an IT system are three separated tasks, but for managers or simple users generating, verifying, and sending an invoice happens in one task (i.e., the invoice is created, verified and sent immediately without any manual intervention). In this case, cp$_{16}$ would be used to merge these three activities into ``Process Invoice'' to better reflect the real-world process.

\textbf{cp$_{17}$} (Delete Entire Branch) removes an entire branch inside gateways with all associated tasks, control edges, and dependencies. The pattern adjusts conditions and flattens the hierarchy by removing gateways if only one branch remains. This pattern improves process refinement and clean-up by reducing errors that might occur if multiple elements had to be deleted one by one. It is particularly useful in GUIs where users cannot easily select multiple elements simultaneously using drag and drop.

\textbf{cp$_{18}$} (Leave Single Branch) removes multiple branches inside gateways with all associated tasks, control edges, and dependencies, leaving only a single branch. The pattern adjusts conditions and flattens the hierarchy by removing gateways as only one branch remains. This pattern improves process refinement and clean-up by reducing errors that might occur if multiple branches had to be deleted one by one. It is particularly useful in GUIs where users cannot easily select multiple elements simultaneously using drag and drop.

\textbf{cp$_{19}$} (Replace Gateways) replaces both splitting and merging components of a gateway simultaneously. This pattern is useful when the control flow behaviour changes (e.g., tasks that were previously executed in parallel are now executed sequentially). Since gateways typically consist of both splitting and merging components, this pattern allows for both parts to be changed at once, rather than modifying them separately to avoid potential inconsistencies.

The list of all existing and potentially required patterns that will be considered further in the paper can be found in Table~\ref{tab:changepatterns}.
These patterns are primarily based on existing literature, practical experience, and assumptions, serving as a solid starting point for understanding user needs in chatbot-related scenarios. Next, we systematically test and evaluate these patterns with real users to assess their effectiveness and relevance in practice.

\begin{table}[!ht]
    \scriptsize
    \centering
    \caption{Overview of the Change Patterns CP (where green - existing Primitives and Patterns, yellow - proposed extension, blue - not considered)}
    \begin{tabular}{p{0.8cm}|p{4.0cm}||p{0.8cm}|p{5.2cm}}
        ID & Name & ID & Name   \\ \hline \hline
        \multicolumn{4}{c}{\textbf{Low-level Primitives}} \\ \hline
        \rowcolor{green!25}
        LP1   & Insert Node & LP2 & Delete Node \\
        \rowcolor{green!25}
        LP3   & Insert Edge & LP4 & Delete Edge \\
        \rowcolor{green!25}
        LP5   & Move Edge & \cellcolor{yellow!25}LP6 & \cellcolor{yellow!25}Rename Node \\  \hline

        \multicolumn{4}{c}{\textbf{High-level Change Patterns}} \\ \hline
        \rowcolor{green!25}
        cp$_1$   & Insert Process Fragment & cp$_2$ & Delete Process Fragment \\
        \rowcolor{green!25}
        cp$_3$   & Move Process Fragment  & cp$_{14}$  & Copy Process Fragment  \\
        \rowcolor{green!25}
        cp$_4$   & Replace Process Fragment  & cp$_5$   & Swap Process Fragments \\
        \rowcolor{yellow!25}
        cp$_{8.1}$ & Embed Process Fragment in Pre-Cond. Loop & cp$_{8.2}$ & Embed Process Fragment in Post-Cond. Loop\\
        \rowcolor{green!25}
        cp$_9$   & Parallelise Process Fragments & cp$_{10}$  & Embed Process Fragment in Cond. Branch \\
        \rowcolor{yellow!25}
        cp$_{15}$  & Split Process Fragment & cp$_{16}$  & Merge Process Fragment  \\ \hline
        \rowcolor{blue!15}
        cp$_{11}$  & Add Control Dependency  &  cp$_{12}$  & Remove Control Dependency  \\ \hline
        \rowcolor{green!25}
        cp$_{13}$  & Update Condition & \cellcolor{yellow!25}cp$_{19}$  & \cellcolor{yellow!25}Replace Gateways \\ \hline
        \rowcolor{green!25}
        cp$_6$   & Extract Sub Process & cp$_7$   & Inline Sub Process \\ \hline
        \rowcolor{yellow!25}
        cp$_{17}$  & Delete Entire Branch & cp$_{18}$  & Leave Single Branch
    \end{tabular}
    \label{tab:changepatterns}
\end{table}

\section{Evaluation Methodology}
\label{sec:eval}

\srmrev{This section describes evaluation concepts and data collection.}
\subsection{Evaluation Concepts}
\label{subsec:validity}

\rnk{CPMR has two goals, i.e., (a) create structurally and semantically correct process models and (b) satisfy user expectations by correctly identifying and implementing their redesign intentions. Thus, the evaluation has to cover both of these aspects, since LLMs running in the background of the proposed approaches can lead to wrong results, and user requirements are inevitably ambiguous or insufficient~\cite{10.1145/3660810}.}

\begin{figure}[htb!]
    \centering
    \includegraphics[width=0.7\textwidth]{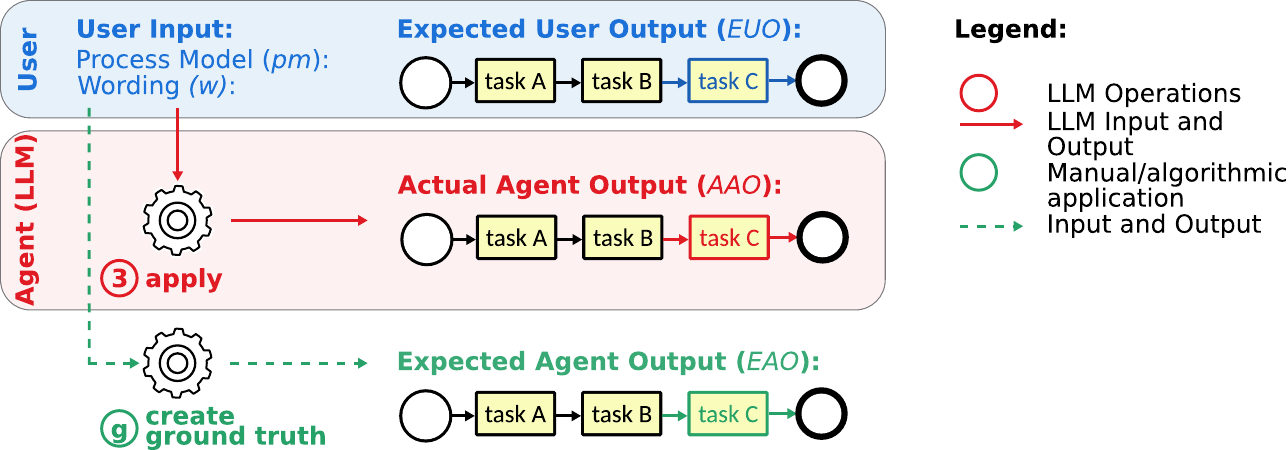}
    \caption{Overview of LLM-based Process Model Redesign}
    \label{fig:eval_concept}
\end{figure}

\srm{The evaluation is thus two-staged. In the first phase, we evaluate the outcomes of functions {\color{red}\ding{172}} identify and {\color{red}\ding{173}} derive as this  provides valuable insights into user behaviour and wording. It helps us understand where users fail to provide meaningful wording, why patterns are sometimes identified but still fail to provide meaning, which aspects of the redesign request might need clearer expression, and whether the existing change patterns are sufficient to cover user behaviour. In the second stage, we evaluate the output of the approach, i.e., the redesigned process model. To this end, we create three versions of the redesigned process model (cf. Fig. \ref{fig:eval_concept}), i.e., the Expected User Output (EUO), the Actual Agent Output (AAO), and the Expected Agent Output (EAO).} EUO is the process model the user intends to obtain. EAO is the new process model representing correct agent behaviour when applying a change pattern after executing function \Circled[inner color=ForestGreen, outer color=ForestGreen]{g} manually or with the algorithm to a given process model. \newnk{In the scope of this work, all EAO models were created manually using a traditional ``drag-and-drop'' modelling approach.}
AAO is the process model created by the agent, i.e., after executing function \Circled[inner color=red, outer color=red]{3} apply (see Fig.~\ref{fig:eval_concept}).

\srm{\noindent\textbf{Stage 1:} At first, we assess the cases in which function {\color{red}\ding{172}} yields $false$ (i.e., no change pattern can be identified based on the user wording) and function {\color{red}\ding{173}} returns $false$ (i.e., change pattern was identified based on the user wording, but it is not possible to derive meaning out of it).
Doing so, we analyse whether the change patterns are properly understood by the agent, potentially identify new patterns in user behaviour that have not previously been defined or considered, and detect discrepancies or errors. Table \ref{tab:validity1} summarises the possible interpretations for $false$ results in functions identify and derive. }

\begin{table}[!ht]
    \centering
    \scriptsize
    \caption{Stage 1: Assessment of Functions identify and derive}
        \begin{tabular}{c|c|p{3.2in}c}
        \textbf{Identify}   & \textbf{Derive} &  \textbf{\makecell[c]{Interpretation}}   \\ \hline
         &  & \textbf{Pattern is not identified:} \\
         &  & \textbf{(a) No match is found.} The system could not map the wording to any existing rule. Potentially, a new change pattern can be identified.  \\
        false  & --  & \textbf{(b) User input is incomplete.} The wording provided by the user lacks sufficient information to determine the intended change. The system should prompt the user for additional details. \\
         &  & \textbf{(c) Multiple matches exist.} The request matches multiple patterns, leading to ambiguity. The system should prompt the user to clarify the request.        \\ \hline
        true & false  &  \textbf{User input is incomplete.} The wording provided by the user lacks sufficient information to determine the intended change (i.e., location, elements, labels, etc.). The system should prompt the user for additional details.        \\
    \end{tabular}
    \label{tab:validity1}
\end{table}

\srm{\noindent\textbf{Stage 2} assumes that functions identify and derive do not result in \emph{false}, but yield a change pattern $cp$ and a meaning $m$ respectively, resulting in AAO after executing function \Circled[inner color=red, outer color=red]{3} apply on process model $pm$. Comparing AAO to EUO enables the assessment of how user expectations are met and comparing AAO to EAO  assesses the effectiveness of LLM-based redesign, i.e.,
the agent performance and the effectiveness of the prompt design. It helps identify patterns that may fail due to incorrect agent interpretation or limitations in the existing pattern descriptions. This evaluation also highlights whether all existing patterns are necessary and which patterns can be considered alternatives. Table~\ref{tab:validity2} summarises interpretations of comparing the redesigned process models AAO, EUO, and EAO. }

\begin{table}[!ht]
    \centering
    \scriptsize
    \caption{Assessing LLM Performance and Effectiveness }
    \begin{tabular}{c|c|p{3in}}
        \textbf{AAO == EUO} & \textbf{AAO == EAO} & \textbf{\makecell[c]{Interpretation}}    \\ \hline
        true           & true        & \textbf{Correct behaviour.} The expected user output matches the actual system output, and the system behaved as expected.      \\ \hline
        true           & false       & \textbf{Incorrect pattern implementation. } The system did not behave as expected even though the user’s request matched an existing pattern and user's expectations.              \\ \hline
         false          & true        & \textbf{Incorrect pattern application or identification. } The user misunderstood how the pattern works and need guidance on proper application or the system mapped wording to a wrong pattern, leading to incorrect changes.       \\ \hline
         false          & false       & \textbf{Critical inconsistency. }The applied pattern produced results that neither match the user's expectations nor align with the expected system behaviour suggesting a fundamental issue with pattern identification, execution or user input. \\
    \end{tabular}
    \label{tab:validity2}
\end{table}

\subsection{Data Collection}
\label{sub:evalprocedure}
User preferences, expectations, and actual usage behaviour may differ substantially from the assumptions that formed the basis of the proposed patterns (see Table~\ref{tab:changepatterns}).
Therefore, it is essential to conduct comprehensive user studies and gather feedback to validate whether these patterns align with user needs and improve the overall user experience. A closer look at how users interact with the system will provide insights for refining these patterns and introducing new ones if necessary. This approach not only helps improve the user experience by enhancing human-computer interaction, but also contributes to formulating a better rule base, making the chatbot more flexible and improving its performance.

\begin{figure}[htb!]
    \centering
    \includegraphics[width=0.6\textwidth]{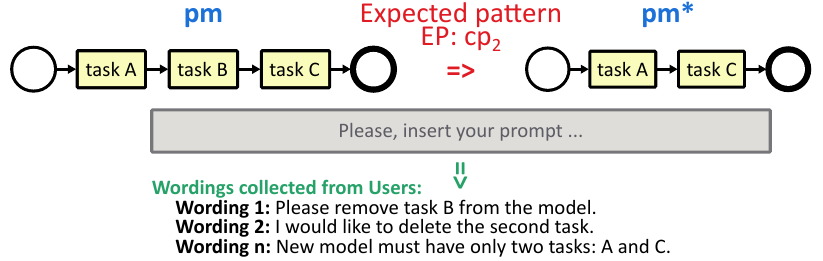}
    \caption{User survey. Example }
    \label{fig:example}
\end{figure}

To analyse user behaviour, we conducted a user survey with 64 participants
\footnote{Survey: \url{https://docs.google.com/forms/d/e/1FAIpQLSerHMjzTR4ll3XJN_vWAZ53KWy1blKLr-fv3Wd0_BvUxUyHWA/viewform}},
\footnote{Survey Overview: \url{https://github.com/com-pot-93/cpd/blob/main/survey/survey-overview.pdf}}. Users were first asked to answer some general questions about their modelling skills, after which they were presented with models pairs ($pm$, $pm^*$). Their task was to imagine that, using a chatbot, they have to make changes to pm to derive $pm^*$, i.e., using natural language to achieve a transformation (see Fig.~\ref{fig:example}).

\begin{table}[!ht]
    \centering
    \scriptsize
    \caption{Statistics about Process Models used in The Survey}
    \begin{tabular}{l|c|c|c|c}
        \textbf{Model} & \textbf{\#Events} & \textbf{\#Tasks}  & \textbf{\#Gateways}  & \textbf{\#Subprocesses}  \\ \hline
        pm           & 2.00	& 4.11	& 1.89	& 0.06 \\
        pm$^*$       & 2.00	& 3.89	& 2.22	& 0.06
    \end{tabular}
    \label{tab:stat}
\end{table}

\rnk{The input process models are simple BPMN models with two events (start and end), up to five tasks, parallel and exclusive gateways, and subprocesses for certain patterns (see Tab.~\ref{tab:stat} for more details)}. The process models were created with dummy tasks to avoid any potential domain biases, especially since some of the participants had no or limited modelling experience. The output models were derived by applying a list of existing and proposed change patterns (see Tab.~\ref{tab:changepatterns}). The list of all input and output process models, as long as the survey results, can be found in\footnote{Input Process Models: \url{https://github.com/com-pot-93/cpd/tree/main/input}}, \footnote{Output Process Models: \url{https://github.com/com-pot-93/cpd/tree/main/EAO}}, \footnote{Survey Data: \url{https://github.com/com-pot-93/cpd/tree/main/survey}}.

Thus the setting of the survey is as follows:

\begin{itemize}
    \item The user is presented with an Input Process Model ($pm$).
    \item The user is presented with an Output Process Model ($pm^*$).
    \item The user is intended to provide a \textit{Wording} that describes how to transform pm into $pm^*$.
    \item The user had to provide a \textit{Wording} for 18 patterns.
    \item For each of the 18 transformations an Expected Pattern (EP) was assigned.
\end{itemize}

The EPs were not communicated to the user. Each user was intended to provide whatever wording they deem necessary. The survey itself covered all patterns, so for each pattern 64 different wordings were collected.

\subsection{Data Evaluation}
\label{sub:evaluation}

\rnk{After collecting the wordings we executed (a) a baseline approach, where the user-provided wording is directly applied (i.e., through an LLM prompt) to a provided process model and (b) the CPMR approach.}

{\nkMulti
\textbf{a) Baseline.} Relying on the experience from our previous work~\cite{coopis}, we adopt the rules defining the desired output format of the redesigned model and the structure of the prompt for process model redesign. This approach is a good starting point for understanding how well the LLM can generate results from direct user input, without any further manipulations.
The prompt for the baseline approach is identical to the \textsl{Apply} prompt (see Sect.~\ref{sub:overview}), with the only difference that instead of passing
the meaning derived from the wording to the LLM, we provide wording directly as a part of the user input.

In the user survey settings, both EAO and EUO are identical, since the users were presented with EAO and their wording describe how they want to achieve this output.
Therefore, the evaluation \newnk{of the baseline} is simplified from presented in Table~\ref{tab:validity2} four states to the final two: either AAO is equal to the EAO (\texttt{True}) or it is not (\texttt{False}) (i.e., pattern success or failure respectively).
}

\textbf{CPMR Approach.} After collecting the wordings the following steps are then performed:

\begin{itemize}
    \item For each wording identify a pattern cp$_x$ if possible.
    \item Compare the pattern cp$_x$ with the expected pattern cp$_y$.
    \item Check if $m$ can be derived successfully from cp$_x$.
    \item When m is applied to pm, check if the resulting pm' is identical with $pm^*$ .
\end{itemize}

The goal is to find out which patterns are useful, which patterns can be replaced, reduced or combined, and if some additional patterns have to be added. For this purpose we follow the procedure, described in Section~\ref{subsec:validity}. For each of the steps \Circled[fill color=mylila1, inner color=black, outer color=black]{1a}, \Circled[fill color=mylila1, inner color=black, outer color=black]{1b}, \Circled[fill color=mylila1, inner color=black, outer color=black]{2}, \Circled[fill color=mylila1, inner color=black, outer color=black]{3a} yields \texttt{True} or \texttt{False}, and map obtained results with the Tables~\ref{tab:validity1} and~\ref{tab:validity2}.

\begin{figure}[htb!]
    \centering
    \includegraphics[width=0.8\textwidth]{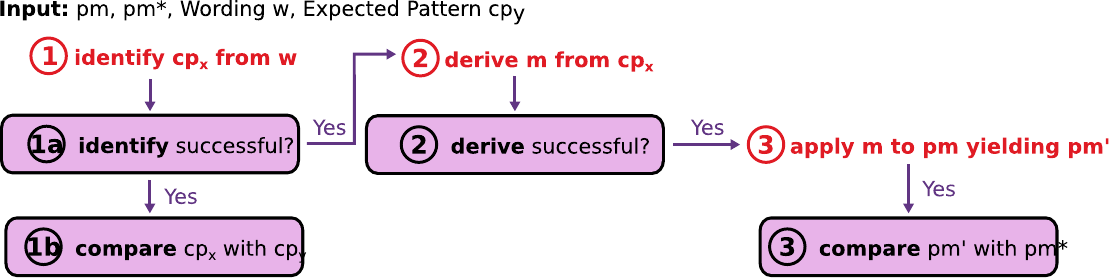}
    \caption{Evaluation Procedure}
    \label{fig:evaluation}
\end{figure}

Figure~\ref{fig:evaluation} presents the agent stages and illustrates how we interpret the output of these stages in order to establish how efficiently the selected LLMs can map the provided wording to the list of existing patterns, whether the mapped patterns are consistent with expectations, whether the provided wording is sufficient for deriving meaning, and whether a new pattern is required. \newnk{Here, even though the EAO and EUO are identical, we still attempt to capture users' expectations by utilizing the identified pattern cp$_x$ along with the expected one cp$_y$. Thus, even without modelling the EUO explicitly, we still refer to the difference between expected user behaviour and actual agent behaviour (i.e., Tab.~\ref{tab:validity2}, AAO == EUO is expressed as
cp$_x$ == cp$_y$).}

\Circled[fill color=mylila1, inner color=black, outer color=black]{1a} The status of the identified pattern cp$_x$ is set to \texttt{True} if it is matched to one of the patterns in CP. If the wording can not not matched to any pattern in the CP, if it matches multiple patterns, or if the agent provides an output different from the \newnk{allowed format (i.e., return not only the pattern itself, but also some side comments, that are not recognised by the system)}, the status is set to \texttt{False}.

\Circled[fill color=mylila1, inner color=black, outer color=black]{1b} If cp$_x$ could be identified, we check whether cp$_x$ equals the expected pattern. If they are identical, the status is set to \texttt{True}; otherwise, the status is set to \texttt{False}.

\Circled[fill color=mylila1, inner color=black, outer color=black]{2}  If cp$_x$ could be identified, we also proceed to the next step and derive the meaning from the wording provided by user. If the meaning is successfully derived (i.e., there is enough information in the wording to perform a change), the status is set to \texttt{True}; otherwise, the status is set to \texttt{False}, and no further steps are performed.

\Circled[fill color=mylila1, inner color=black, outer color=black]{3} If the meaning is successfully derived, we apply it to pm* and generate the AAO as pm', which is then compared to the pm* from the survey (which we define to be the EAO).

In the end, we have six possible combinations introduced in Table~\ref{tab:validity1} and~\ref{tab:validity2} to evaluate how sufficient the selected change patterns are and, at the same time, how efficient the LLMs are in process model redesign. Only the option in which all four steps return \texttt{True} is considered successful, and all others are classified as failures.

\textbf{Note.} In both cases, to compare two process models, we use element-based semantic similarity, first introduced in~\cite{voelter2024}. Each element of the process model is considered, and for each element, semantically similar elements in the second model are identified using the \textit{dice score}. The overall score is then weighted by the harmonic mean of the lengths of the elements. The similarity value ranges from 0 to 1.

Since the models in the survey are very simple, and even small differences result in different models, we set the threshold to 1. Thus, if the similarity between the AAO and the EAO is equal to 1, the status is set to \texttt{True}; otherwise, the status is set to \texttt{False}.

\nk{Such a strict threshold ensures that the models are not only structurally identical but also equivalent in terms of completeness (i.e., all elements in one process model are included in the other) and correctness (i.e., the tasks not only have matching labels but also appear in the same sequence). In doing so, it also serves as an indicator of user satisfaction with the outcome, as only fully accurate and complete models are considered equivalent.}

\section{Results}
\label{sec:evaluation}

For the evaluation, we use three different large language models (LLMs), i.e., \texttt{gpt-4o}, \texttt{gemini-2.0-flash}, and \texttt{mistral-large-latest} (hereafter referred to as gpt, gemini, and mistral, respectively)\footnote{\small \rnk{We provide the IDs of the LLMs used in the API, which always rely on the latest deployed version of the model (last used: 17–24 November 2025).}}.
\rnk{All three LLMs can be classified as high-performance models and each of them has advantages in different aspects that make it particularly suitable for the model redesign task. }

{\nkMulti
\texttt{gpt-4o} enables real-time interactions without losing performance for complex, reasoning-heavy tasks, making it appropriate for process model redesign tasks. \texttt{gemini-2.0-flash} offers a balance between cost and performance. With its fast response times, low resource usage, and ability to handle large inputs, this LLM is suitable for use cases where quick feedback is required, making it a good candidate for tasks like process model adjustment. Similar to \texttt{gemini-2.0-flash}, \texttt{mistral-large-latest} is also a good choice for tasks requiring real-time interaction and is more flexible regarding the deployed and integrated it in customised environments.

Open-source models are not considered in this user study, as large models with a higher number of parameters (e.g., llama3.1 70b-405b, deepseek-v3.1 671b, etc.) require specialised hardware for efficient computation. Attempts with smaller models, such as phi4-14b, llama3.1 8b, deepseek-r1 7-32b, mixtral 8x7b, etc., achieve poor non-meaningful results and therefore are not included in the evaluation.
}

Based on the results obtained from~\cite{coopis}, we adopt the strategy of addressing tasks by their labels, performing only one change at a time, and passing prompt-related information to the LLM (input process model, change to be performed, and rules for the output format). We also use a zero-shot strategy and provide no examples of process descriptions, input, or output process models (see Section~\ref{sub:overview} for more details). The prompts used during the evaluation along with all generated artifacts and non-average data are available in\footnote{Prompts: \url{https://github.com/com-pot-93/cpd/tree/main/prompts}}\footnote{Generated Data: \url{https://github.com/com-pot-93/cpd/tree/main}}.

{\nkMulti
\subsection{Baseline Approach}

We consider cases where the actual and expected agent outputs (AAO and EAO) are equally correct (see Sect.~\ref{sub:evaluation}). Table~\ref{tab:baseline} presents the average results for three selected LLMs for all change patterns. Even though the approach itself does not require change patterns, we sort all the results according to the patterns to which the wording in the survey was originally provided.

\begin{table}[!ht]
    \centering
    \scriptsize
    \caption{Correct Behaviour across Change Patterns: \newnk{Values represent the percentage} of cases in which EAO equals AAO}
    \begin{tabular}{l|c|c|c|c|c|c|c|c|c}
         Method & \textbf{cp$_1$} & \textbf{cp$_2$} & cp$_3$ & \textbf{cp$_4$} & \textbf{cp$_5$} & cp$_6$ & cp$_7$ & cp$_{8.1}$ & cp$_{8.2}$ \\ \hline \hline
        gemini & 0.86 & 0.92 & 0.31 & 0.92 & 0.86 & 0.31 & 0.25 & 0.02 & 0.06 \\
        gpt & 0.91 & 0.86 & 0.45 & 0.94 & 0.92 & 0.58 & 0.36 & 0.08 & 0.14 \\
        mistral & 0.92 & 0.59 & 0.44 & 0.94 & 0.92 & 0.31 & 0.27 & 0.08 & 0.08 \\ \hline
        \rowcolor{green!25}
        average & \textbf{0.90} & \textbf{0.79} & 0.40 & \textbf{0.93} & \textbf{0.90} & 0.40 & 0.29 & 0.06 & 0.09 \\
        \multicolumn{10}{c}{~} \\
        Method & cp$_9$ & cp$_{10}$ & \textbf{cp$_{13}$} & \textbf{cp$_{14}$} & \textbf{cp$_{15}$} & \textbf{cp$_{16}$} & \textbf{cp$_{17}$} & cp$_{18}$ & \textbf{cp$_{19}$} \\ \hline \hline
        gemini & 0.20 & 0.05 & 0.98 & 0.58 & 0.84 & 0.50 & 0.73 & 0.39 & 0.67 \\
        gpt & 0.17 & 0.03 & 1.00 & 0.64 & 0.88 & 0.47 & 0.80 & 0.44 & 0.83 \\
        mistral & 0.13 & 0.13 & 0.97 & 0.53 & 0.92 & 0.55 & 0.84 & 0.50 & 0.77 \\ \hline
        \rowcolor{green!25}
        average & 0.17 & 0.07 & \textbf{0.98} & \textbf{0.58} & \textbf{0.88} & \textbf{0.51} & \textbf{0.79} & 0.44 & \textbf{0.76} \\
    \end{tabular}
    \label{tab:baseline}
\end{table}

Based on the results presented in the Table~\ref{tab:baseline} we can see, that 10 out of 18 change patterns perform good (cp1, cp2, cp4, cp5, cp13, cp14, cp15, cp16, cp17, and cp19), achieving accuracy of 50\% or more, where 8 of these 10 patterns achieving between 76\% and 98\% accuracy on average (cp1, cp2, cp4, cp5, cp13, cp15, cp17, and cp19).

Eight patterns (cp3, cp6, cp7, cp8.1, cp8.2, cp9, cp10, and cp18) stay under the 50\% accuracy threshold, sometimes reaching only between 6\% and 7\% accuracy on average.

\newnk{In 10 of the 18 patterns, the gpt model performs best. In 12 of the 18 patterns, gemini shows the worst performance. However, for the majority of patterns, the performance range between the LLMs is relatively small, i.e., between 2\% and 14\%-points. Only patterns cp2, cp6, and cp19 differ significantly, with performance ranges of 33\%, 27\%, and 16\%, respectively.}

It is important to mention that not only well-established patterns succeed, but also patterns that were newly proposed in Section~\ref{sec:patterns}.
}

\subsection{CPMR Approach}

As a starting point, we consider cases where both the user and the agent perform well, and the changes provided by the user were successfully implemented, \rnk{i.e., for one particular wording each step described in Section~\ref{sub:evaluation} (\Circled[fill color=mylila1, inner color=black, outer color=black]{1a}, \Circled[fill color=mylila1, inner color=black, outer color=black]{1b}, \Circled[fill color=mylila1, inner color=black, outer color=black]{2}, and \Circled[fill color=mylila1, inner color=black, outer color=black]{3}) returns the status \texttt{True}. In comparison to the baseline approach, we define a pattern as successfully implemented if it reaches at least 30\% (and not 50\%), since we have filtering mechanisms that prevent us from counting the identical process models $pm'$ and $pm*$. }

As can be seen in Table~\ref{tab:matching-existing}, only 9 out of 18 patterns were realised correctly in more than 30\% of all cases. For the already well-established change patterns (cp$_1$–cp$_{14}$, excluding pattern cp$_8$), patterns cp$_1$, cp$_2$, cp$_4$, cp$_5$, and cp$_{13}$ reach the 30\% threshold. For the proposed patterns (cp$_{15}$–cp$_{19}$, cp$_{8.1}$, and cp$_{8.2}$), four patterns (i.e., cp$_{15}$–cp$_{17}$, and cp$_{19}$) succeeded.

\begin{table}[!ht]
    \centering
    \scriptsize
    \caption{Correct Behaviour across Change Patterns: \newnk{Values represent the percentage} of cases in which EAO equals AAO and cp$_x$ equals cp$_y$}
    \begin{tabular}{l|c|c|c|c|c|c|c|c|c}
        Method & \textbf{cp$_1$} & \textbf{cp$_2$} & cp$_3$ & \textbf{cp$_4$} & \textbf{cp$_5$} & cp$_6$ & cp$_7$ & cp$_{8.1}$ & cp$_{8.2}$ \\ \hline \hline
        gemini & 0.69 & 0.64 & 0.16 & 0.63 & 0.53 & 0.00 & 0.02 & 0.00 & 0.02 \\
        gpt & 0.73 & 0.56 & 0.27 & 0.63 & 0.84 & 0.20 & 0.08 & 0.05 & 0.02 \\
        mistral & 0.75 & 0.17 & 0.36 & 0.53 & 0.77 & 0.05 & 0.05 & 0.02 & 0.03 \\ \hline
        \rowcolor{green!25}
        average & \textbf{0.72} & \textbf{0.46} & 0.26 & \textbf{0.59} & \textbf{0.71} & 0.08 & 0.05 & 0.02 & 0.02 \\
        \multicolumn{10}{c}{~} \\
        Method & cp$_9$ & cp$_{10}$ & \textbf{cp$_{13}$} & cp$_{14}$ & \textbf{cp$_{15}$} & \textbf{cp$_{16}$} & \textbf{cp$_{17}$} & cp$_{18}$ & \textbf{cp$_{19}$} \\ \hline \hline
        gemini & 0.02 & 0.00 & 0.84 & 0.00 & 0.53 & 0.27 & 0.64 & 0.00 & 0.25 \\
        gpt & 0.13 & 0.03 & 0.91 & 0.02 & 0.59 & 0.38 & 0.69 & 0.09 & 0.61 \\
        mistral & 0.13 & 0.03 & 0.91 & 0.00 & 0.58 & 0.27 & 0.67 & 0.02 & 0.52 \\ \hline
        \rowcolor{green!25}
        average & 0.09 & 0.02 & \textbf{0.89} & 0.01 & \textbf{0.57} & \textbf{0.30} & \textbf{0.67} & 0.04 & \textbf{0.46} \\
    \end{tabular}
    \label{tab:matching-existing}
\end{table}

\rnk{
In Tables~\ref{tab:baseline} and~\ref{tab:matching-existing}, we can see that $8$ patterns are systematically underperforming. These patterns are cp$_3$, cp$_6$, cp$_7$, cp$_{8.1}$, cp$_{8.2}$, cp$_9$, cp$_{10}$, and cp$_{18}$. There is also one pattern (i.e., cp$_{14}$) that performs well for the baseline, but not using the CPMR approach.} To understand why these patterns fall below threshold, we examine more closely the cases where either identification was not possible, or it was not possible to derive a meaning from the provided wording (Stage 1, see Sect.~\ref{subsec:validity}), or where the wording and the meaning were considered sufficient by the agent but the user applied the wrong pattern, or where the agent implemented the pattern incorrectly (Stage 2, see Sect.~\ref{subsec:validity}).

\textbf{Pattern is not identified.} First, we analysed all cases where the identification received a \texttt{False} status: the pattern was not identified or the identified pattern did not match any existing pattern in CP \rnk{(i.e., \Circled[fill color=mylila1, inner color=black, outer color=black]{1a} is \texttt{False} for each wording, see Section~\ref{sub:evaluation} for more details).}
According to Table~\ref{tab:validity1}, there are three possible reasons for such behaviour, and only in one of these cases a new pattern can be required.

In Table~\ref{tab:mapping-false}, we can see that, generally, for each of the patterns in CP, between 3\% and 25\% of the cases had wording that was not matched to existing patterns. In detail, we will consider only the patterns that reach at least 10\% of unidentified cases, e.g., patterns \textbf{cp$_4$, cp$_7$, cp$_{10}$, cp$_{16}$, and cp$_{18}$}.

\begin{table}[!ht]
    \centering
    \scriptsize
    \caption{Unidentified Change Patterns: \newnk{Values represent the} percentage of cases in which cp$_x$ was not defined from $w$}
    \renewcommand{\tabularxcolumn}[1]{m{#1}}
    \begin{tabular}{l|c|c|c|c|c|c|c|c|c}
        Method & cp$_1$ & cp$_2$ & cp$_3$ & \textbf{cp$_4$} & cp$_5$ & cp$_6$ & \textbf{cp$_7$} & cp$_{8.1}$ & cp$_{8.2}$ \\ \hline \hline
        gemini & 0.02 & 0.03 & 0.06 & 0.20 & 0.09 & 0.16 & 0.16 & 0.08 & 0.05 \\
        gpt & 0.02 & 0.03 & 0.02 & 0.22 & 0.03 & 0.05 & 0.06 & 0.06 & 0.05 \\
        mistral & 0.05 & 0.03 & 0.05 & 0.33 & 0.09 & 0.08 & 0.13 & 0.05 & 0.09 \\ \hline
        \rowcolor{green!25}
        average & 0.03 & 0.03 & 0.04 & \textbf{0.25} & 0.07 & 0.09 & \textbf{0.11} & 0.06 & 0.06 \\
        \multicolumn{10}{c}{~} \\
        Method & cp$_9$ & \textbf{cp$_{10}$} & cp$_{13}$ & cp$_{14}$ & cp$_{15}$ & \textbf{cp$_{16}$} & cp$_{17}$ & \textbf{cp$_{18}$} & cp$_{19}$ \\ \hline \hline
        gemini & 0.11 & 0.08 & 0.05 & 0.08 & 0.08 & 0.09 & 0.05 & 0.11 & 0.08 \\
        gpt & 0.06 & 0.09 & 0.05 & 0.08 & 0.08 & 0.08 & 0.05 & 0.11 & 0.08 \\
        mistral & 0.06 & 0.13 & 0.06 & 0.08 & 0.09 & 0.17 & 0.09 & 0.11 & 0.11 \\ \hline
        \rowcolor{green!25}
        average & 0.08 & \textbf{0.10} & 0.05 & 0.08 & 0.08 & \textbf{0.11} & 0.06 & \textbf{0.11} & 0.09 \\
    \end{tabular}
    \label{tab:mapping-false}
\end{table}

In most of these cases, the user input (i.e., wording) did not align with any existing pattern, because the user was unable to describe the transformation from the input process model pm to the output process model $pm*$ (e.g.,``I don’t know'', empty wording, or irrelevant content).

For pattern \textbf{cp$_4$}, users often referred to an operation that is not defined as a pattern but as a proposed low-level primitive — ``Rename node'' (e.g., in the provided example: renaming a label or task name). Its utilization and frequency of occurrence suggest that extending the existing set of low-level primitives to include this action is appropriate.

In the case of patterns \textbf{cp$_7$}, \textbf{cp$_{10}$} and \textbf{cp$_{18}$}, the user was not referring to the patterns themselves, but \newnk{referred to a previous question (e.g., ``reverse the action from the previous question'') making the wording non self-contained, as the referenced information was not available to the LLM. Alternatively, in some cases, the user requested to delete/ignore the provided process model and create a new process from scratch, which is not aligned with the task assigned to the LLM.}

For pattern \textbf{cp$_{16}$}, in multiple cases, instead of using common terms like merge, combine, or join, users chose to describe the transformation using the term ``summarise'' (e.g.,  ``Summarize task B and task E into a single task B\&E.''). While this can be interpreted as cp$_{16}$, it is not immediately clear, since the initial meaning of summarizing is \textit{``to express the most important facts or ideas about something or someone in a short and clear form''}\footnote{\url{https://dictionary.cambridge.org/dictionary/english/summarizing}}.

\textbf{Meaning is not derived.}
Next, we examine the cases in which the pattern was identified, but the meaning was not derived \rnk{(i.e., \Circled[fill color=mylila1, inner color=black, outer color=black]{1a} is \texttt{True} but \Circled[fill color=mylila1, inner color=black, outer color=black]{2} is \texttt{False} for each wording, see Section~\ref{sub:evaluation} for more details).} On average, up to 10\% of all cases in all patterns failed to derive a meaning (see Tabl.~\ref{tab:no-meaning}). In most cases, the failure was due to the incorrect wording, which was insufficient to capture the proper meaning of the user's request, \rnk{even though it was clear which pattern the user was referencing}.
\newnk{For instance, the user describes the kind of change they are willing to perform, but the details are not specified. Thus, the request does not contain sufficient information to be performed: e.g., ``Add task E'' or ``Removing a task'' specifies the type of action to be done, but it is not explained how/where it has to be done in the model precisely.}

\newnk{Several other errors might occur. Users frequently describe process changes using informal or incorrect terminology that is not directly interpretable in terms of BPMN semantics such as false task, false connection, or confusing OR and XOR gateways.
Examples of requests are ``Combine task B and E into one process'', ``remove option c'', ``Delete all the judgements'', ``Replace multiplications with additions'', ``Task C is removed, meaning a false task A, won't execute a task C anymore'', and ``Please change the true and false regulation ...''.}

\newnk{Some requests describe the desired execution behaviour rather than concrete model transformations, e.g., ``Instead of being executed at the same time, b and e are now executed sequentially''. While execution as a concept is relevant for BPMN models, such statements are difficult to directly map to structural model edits and hence not directly interpretable for a redesign agent. }

\newnk{User input may contain inconsistencies, resulting in contradictory instructions, e.g., ``Split task B \& E in two separated tasks called task B and task E. And replace the two new tasks with the task B \& E.'', ``Task C is removed, and a task C is added. [...]'', ``Please remove task D from its current location. [...] Connect the two new if-else blocks with task D''.}

\newnk{In some cases, failures are caused by instructions that modify BPMN elements without specifying how to update their surrounding elements, making the requested change ambiguous, e.g., ``Exclusive gateway was removed''.}

\newnk{The result of such wording is that the agent is unable to match the patterns properly. However, there are also situations where the current implementation cannot fully capture and preprocess the user’s intent. For instance, when a user, instead of referring to a single pattern, tries to utilise multiple patterns, it is a valid action (since most patterns can be realised through other patterns), e.g., ``Add task B and Task E after task B\&E. Remove task B\&E'' or ``Delete task E. Rename task B to task B\&E.''. This indicates that the system must be able to recognise and implement multiple pattern utilization.}

\begin{table}[!ht]
    \centering
    \scriptsize
    \caption{Change Patterns without Meaning: \newnk{Values represent the} percentage of cases in which $m$ was not derived from cp$_x$}
    \renewcommand{\tabularxcolumn}[1]{m{#1}}
    \begin{tabular}{l|c|c|c|c|c|c|c|c|c}
        Method & cp1 & cp2 & cp3 & cp4 & \textbf{cp5} & cp6 & cp7 & \textbf{cp8.1} & \textbf{cp8.2} \\ \hline  \hline
        gemini & 0.11 & 0.06 & 0.06 & 0.06 & 0.31 & 0.17 & 0.13 & 0.22 & 0.19 \\
        gpt & 0.03 & 0.02 & 0.14 & 0.02 & 0.06 & 0.05 & 0.11 & 0.05 & 0.08 \\
        mistral & 0.00 & 0.08 & 0.06 & 0.00 & 0.03 & 0.03 & 0.05 & 0.03 & 0.09 \\ \hline
        \rowcolor{green!25}
        average & 0.05 & 0.05 & 0.09 & 0.03 & \textbf{0.14} & 0.08 & 0.09 & \textbf{0.10} & \textbf{0.12} \\
        \multicolumn{10}{c}{~} \\
        Method & \textbf{cp9} & cp10 & cp13 & cp14 & \textbf{cp15} & \textbf{cp16} & \textbf{cp17} & \textbf{cp18} & \textbf{cp19} \\ \hline  \hline
        gemini & 0.25 & 0.08 & 0.06 & 0.16 & 0.28 & 0.20 & 0.16 & 0.22 & 0.48 \\
        gpt & 0.20 & 0.03 & 0.00 & 0.09 & 0.17 & 0.08 & 0.08 & 0.17 & 0.09 \\
        mistral & 0.13 & 0.06 & 0.00 & 0.06 & 0.05 & 0.16 & 0.06 & 0.08 & 0.00 \\  \hline
        \rowcolor{green!25}
        average & \textbf{0.19} & 0.06 & 0.02 & 0.10 & \textbf{0.17} & \textbf{0.15} & \textbf{0.10} & \textbf{0.16} & \textbf{0.19} \\
    \end{tabular}
    \label{tab:no-meaning}
\end{table}

\rnk{The presence of non-derived meanings for all LLMs across all patterns demonstrate that (a) insufficient wordings can be identified and (b) if LLMs are not pushed to immediately generate an output (e.g., apply wording directly to derive a new process model), but have a chance to identify inaccurate content, they can return this information as feedback that can be used to refine and improve the wording before generating the final output.}

\textbf{Incorrect Pattern Application or Identification.} In this case, the user \newnk{provides the intended changes using} a pattern different from the expected one but still got the same result as expected \rnk{( i.e., for one particular wording steps \Circled[fill color=mylila1, inner color=black, outer color=black]{1a}, \Circled[fill color=mylila1, inner color=black, outer color=black]{2}, and \Circled[fill color=mylila1, inner color=black, outer color=black]{3} return status \texttt{True}, but \Circled[fill color=mylila1, inner color=black, outer color=black]{1b} is equal to \texttt{False})}. In this situation, we can talk either about pattern alternatives, i.e., where one pattern can be used to realise another, either consistently or in some particular scenario, or about agent pattern misidentification, i.e., when the user's wording is correct, but for some reason, the agent matches the wording to the wrong pattern.

To do so, we further examine which other pattern appears most frequently in those misidentification cases, i.e., we look for the predominant alternative pattern. This helps us develop an intuition about how users interpret and understand patterns, and identify which patterns tend to co-occur, serve as alternatives, or are conceptually related to the expected ones.

\begin{table}[!ht]
    \centering
    \scriptsize
    \caption{Incorrect Pattern Application or Identification across Change Patterns: \newnk{Values represent the} percentage of cases in which cp$_x$ and $m$ are defined, EAO equals AAO, but cp$_x$ is not equal to cp$_y$}
    \renewcommand{\tabularxcolumn}[1]{m{#1}}
    \begin{tabular}{l|c|c|c|c|c|c|c|c|c}
        Method & cp$_1$ & \textbf{cp$_2$} & cp$_3$ & cp$_4$ & cp$_5$ & \textbf{cp$_6$} & \textbf{cp$_7$} & cp$_{8.1}$ & cp$_{8.2}$ \\ \hline \hline
        gemini & 0.08 & 0.16 & 0.00 & 0.05 & 0.00 & 0.17 & 0.17 & 0.02 & 0.03 \\
        gpt & 0.08 & 0.27 & 0.02 & 0.08 & 0.02 & 0.25 & 0.22 & 0.02 & 0.02 \\
        mistral & 0.09 & 0.27 & 0.05 & 0.03 & 0.02 & 0.05 & 0.08 & 0.00 & 0.00 \\ \hline
       \rowcolor{green!25}
        average & 0.08 & \textbf{0.23} & 0.02 & 0.05 & 0.01 & \textbf{0.16} & \textbf{0.16} & 0.01 & 0.02 \\  \hline
        \rowcolor{green!25}
        \makecell{predominant\\pattern} & - & cp$_{17}$ & - & - & - & cp$_4$ & \makecell{cp$_4$,\\cp$_{10}$} & - & - \\
        \multicolumn{10}{c}{~} \\
        Method & cp$_9$ & cp$_{10}$ & cp$_{13}$ & \textbf{cp$_{14}$} & cp$_{15}$ & cp$_{16}$ & cp$_{17}$ & \textbf{cp$_{18}$} & cp$_{19}$ \\ \hline  \hline
        gemini & 0.00 & 0.02 & 0.05 & 0.27 & 0.06 & 0.02 & 0.08 & 0.17 & 0.03 \\
        gpt & 0.00 & 0.00 & 0.05 & 0.53 & 0.05 & 0.03 & 0.05 & 0.27 & 0.09 \\
        mistral & 0.02 & 0.00 & 0.03 & 0.45 & 0.06 & 0.05 & 0.09 & 0.25 & 0.14 \\ \hline
        \rowcolor{green!25}
        average & 0.01 & 0.01 & 0.04 & \textbf{0.42} & 0.06 & 0.03 & 0.07 & \textbf{0.23} & 0.09 \\ \hline
        \rowcolor{green!25}
        \makecell{predominant\\pattern} & - & - & - & cp$_1$ & - & - & - & cp$_{17}$ & - \\
    \end{tabular}
    \label{tab:mismatching}
\end{table}

In Table~\ref{tab:mismatching}, we present a summary of the results for the cases where the wording was mapped to the wrong pattern, as well as the pattern to which it was mapped to (i.e., see the predominant pattern in Table~\ref{tab:mismatching}) in cases where, in more than 10\% of instances, all three LLMs consistently predict one or two alternate patterns that differ from the expected one. In other cases, further examination is not conducted, as we consider such pattern variation to be random and not significant.

For each of these predominant patterns, we assess whether the issue arises from user wording (i.e., misunderstanding of the pattern) or incorrect reasoning by the LLM (see Tab.~\ref{tab:alternative}).

Wrong user wording \textbf{(User)} means that, in this case, the application of the provided wording cannot result in the expected output process model or violates BPMN 2.0 rules. Wrong LLM reasoning \textbf{(LLM)} means that the user's wording was correct, but its understanding by the LLM is inconsistent with the true pattern definition and thus needs to be adjusted.

In all cases, we provide an example of user wording in Table~\ref{tab:alternative}. Additionally, when the likely cause of the mismatch is an LLM error or pattern ambiguity, we offer suggestions for why the LLM might fail to match the expected pattern or where clarifying the pattern definitions may help guide both, LLM reasoning and user interpretation more effectively.

{
\scriptsize
\begin{xltabular}{155pt}{>{\scriptsize}c|>{\scriptsize}c|>{\scriptsize}c|>{\scriptsize}c|>{\scriptsize}X}
    \caption{Incorrect Pattern Application or Identification: Detailed Analysis}
    \scriptsize
    \label{tab:alternative} \\
    Pattern & \makecell{Misclas-\\sified As} & \makecell{Inter-\\changeable?} & Cause & Example and Explanation \\ \hline
    \endfirsthead
    \multicolumn{5}{c}%
    {\tablename\ \thetable{}: Incorrect Pattern Application or Identification: Detailed Analysis} \\
    Pattern & \makecell{Misclas-\\sified As} & \makecell{Inter-\\changeable?} & Cause & Example and Explanation \\ \hline
    \endhead
        cp$_2$ & cp$_{17}$ & No & LLM & \textit{``Remove task C such that no task is executed in the false-branch.''} \\
        ~ & ~ & ~ & ~ & This result is highly dependent with the selected example, where the user's intent is to delete a specific task, which would result in the false branch being empty. \\ \hline

        cp$_6$ & cp$_4$ & No & LLM, User  & \textit{``After 'task A' and before the 'end place', replace all elements with a transition 'subprocess F'.''} \\
        ~ & ~ & ~ & ~ & The distinction between "extracting a subprocess" and "replacing elements with one" is subtle and often blurred. This suggests that clearer definitions and separation between cp$_4$ and cp$_6$ are necessary to avoid such ambiguity in both user understanding and model classification. \\ \hline
        cp$_7$ & cp$_4$ & No &  LLM, User & \textit{``Replace subprocess F with the following process: A conditional branch where the 'true' branch contains task B and the 'false' branch contains task C, after the branches converge task D.''} \\
        ~ & ~ & ~ & ~ & The distinction between "replacing a subprocess with elements" and "inlining a subprocess" is often not clear. The clearer definitions and separation between cp$_4$ and cp$_7$ are necessary to avoid such ambiguity in both user understanding and LLM classification.\\ \hline
        cp$_7$ & cp$_{10}$ & No & LLM  &\textit{ ``Expand subprocess F and execute B, if task A returns true, or C if it returns false. After both cases, execute D.''} \\
        ~ & ~ & ~ & ~ & The misclassification is caused by the LLM overemphasising the conditional logic in the prompt, incorrectly interpreting a clear subprocess inlining as a conditional insertion. \\ \hline
        cp$_{14}$ & cp$_1$ & Yes & User  & \textit{``Insert task A' before task D.'' }\\ \hline
        cp$_{18}$ & cp$_{17}$ & Yes & User  & \textit{``Delete the branches with condition b and with condition c.''} \\
    \end{xltabular}
}
\pagebreak
\textbf{Incorrect Pattern Implementation.}
Here, we examine how often the agent fails to perform the changes correctly, even though the wording was complete, the correct pattern was identified, and the meaning was extracted \rnk{(i.e., for one particular wording, steps \Circled[fill color=mylila1, inner color=black, outer color=black]{1a}, \Circled[fill color=mylila1, inner color=black, outer color=black]{1b}, and \Circled[fill color=mylila1, inner color=black, outer color=black]{2} return status \texttt{True}, but \Circled[fill color=mylila1, inner color=black, outer color=black]{3} is equal to \texttt{False})}.

As can be seen, there are many patterns that exceed the 10\% threshold, reaching failure rates as high as 70\%. This indicates that our agent, first, cannot easily handle more complicated patterns. Second, suffers from issues with meaning extraction from the wording. Third, that the high threshold we set for model comparison may be too severe for some cases, which should be examined in more detail.

A detailed examination of these cases can help to determine whether adjustments to the threshold, the prompt itself, or the logic and sequence of the steps (e.g., the analysis of user input, the extraction of meaning from that input, and its application to adjust the initial process model) are necessary.

\begin{table}[!ht]
    \centering
    \scriptsize
    \caption{Incorrect Pattern Implementation across Change Patterns: \newnk{Values represent the} percentage of cases in which cp$_x$ and $m$ are defined, cp$_x$ equals cp$_y$, but EAO is not equal to AAO}
    \renewcommand{\tabularxcolumn}[1]{m{#1}}
    \begin{tabular}{l|c|c|c|c|c|c|c|c|c}
        Method & cp$_1$ & cp$_2$ & \textbf{cp$_3$} & cp$_4$ & cp$_5$ & \textbf{cp$_6$} & \textbf{cp$_7$} & \textbf{cp$_{8.1}$} & \textbf{cp$_{8.2}$} \\ \hline \hline
        gemini & 0.03 & 0.00 & 0.56 & 0.05 & 0.02 & 0.34 & 0.17 & 0.31 & 0.58 \\
        gpt & 0.03 & 0.05 & 0.44 & 0.05 & 0.02 & 0.34 & 0.28 & 0.27 & 0.50 \\
        mistral & 0.05 & 0.20 & 0.30 & 0.06 & 0.05 & 0.66 & 0.27 & 0.16 & 0.56 \\ \hline
        \rowcolor{green!25}
        average & 0.04 & 0.08 & \textbf{0.43} & 0.05 & 0.03 & \textbf{0.45} & \textbf{0.24} & \textbf{0.24} & \textbf{0.55} \\
        \multicolumn{10}{c}{~} \\
        Method & \textbf{cp$_9$} & \textbf{cp$_{10}$} & cp$_{13}$ & \textbf{cp$_{14}$} & cp$_{15}$ & \textbf{cp$_{16}$} & cp$_{17}$ & cp$_{18}$ & cp$_{19}$ \\ \hline \hline
        gemini & 0.47 & 0.72 & 0.00 & 0.11 & 0.03 & 0.27 & 0.00 & 0.13 & 0.05 \\
        gpt & 0.31 & 0.75 & 0.00 & 0.11 & 0.09 & 0.28 & 0.00 & 0.08 & 0.02 \\
        mistral & 0.39 & 0.63 & 0.00 & 0.09 & 0.08 & 0.14 & 0.02 & 0.05 & 0.11 \\ \hline
        \rowcolor{green!25}
        average & \textbf{0.39} & \textbf{0.70} & 0.00 & \textbf{0.10} & 0.07 & \textbf{0.23} & 0.01 & 0.08 & 0.06 \\
    \end{tabular}
    \label{tab:tttf}
\end{table}

\newnk{For these patterns above the 10\% threshold, it can be observed that even though failure rates vary substantially between LLMs for a particular pattern, all three LLMs consistently fail on the same patterns. This indicates that LLMs share a common knowledge base about process models, change patterns, and BPMN notation, regardless of their version. }

\textbf{Critical Inconsistency.} Here, we examine how often the agent fails to perform the changes correctly, even though the wording was complete and the meaning was derived \rnk{(i.e., for one particular wording, steps \Circled[fill color=mylila1, inner color=black, outer color=black]{1a} and  \Circled[fill color=mylila1, inner color=black, outer color=black]{2} are  \texttt{True}, but \Circled[fill color=mylila1, inner color=black, outer color=black]{1b} and \Circled[fill color=mylila1, inner color=black, outer color=black]{3} are \texttt{False})}. However, in comparison with the \textbf{Incorrect Pattern Implementation}, we cannot be sure who or what is responsible for the failure: the user or the LLM.

Looking at Table~\ref{tab:ttff}, we can see that many patterns failed, reaching up to 56\% failure. The analysis of patterns~\textbf{cp$_2$, cp$_6$, cp$_7$, cp$_{14}$, and cp$_{18}$} can be found in \textbf{Incorrect Pattern Application or Identification} (see Tab.~\ref{tab:alternative}).

\begin{table}[!ht]
    \centering
    \scriptsize
    \caption{Critical Inconsistency across Change Patterns: \newnk{Values represent the} percentage of cases in which cp$_x$ and $m$ are defined, but cp$_x$ is not equal to cp$_y$ and EAO is not equal to AAO}
    \renewcommand{\tabularxcolumn}[1]{m{#1}}
    \begin{tabular}{l|c|c|c|c|c|c|c|c|c}
        Method & cp$_1$ & \textbf{cp$_2$} & \textbf{cp$_3$} & cp$_4$ & cp$_5$ & \textbf{cp$_6$} & \textbf{cp$_7$} & \textbf{cp$_{8.1}$} & \textbf{cp$_{8.2}$} \\ \hline \hline
        gemini & 0.08 & 0.11 & 0.16 & 0.02 & 0.05 & 0.16 & 0.36 & 0.38 & 0.14 \\
        gpt & 0.11 & 0.08 & 0.13 & 0.02 & 0.03 & 0.11 & 0.25 & 0.56 & 0.34 \\
        mistral & 0.06 & 0.25 & 0.19 & 0.05 & 0.05 & 0.14 & 0.44 & 0.75 & 0.22 \\ \hline
        \rowcolor{green!25}
        average & 0.08 & \textbf{0.15} & \textbf{0.16} & 0.03 & 0.04 & \textbf{0.14} & \textbf{0.35} & \textbf{0.56} & \textbf{0.23} \\ \hline
        \rowcolor{green!25}
        \makecell{predominant\\pattern} & - & cp$_{17}$ & cp$_{10}$ & - & - & cp$_4$ & \makecell{cp$_4$,\\cp$_{10}$} & \makecell{cp$_{8.2}$,\\cp$_{10}$} & cp$_{10}$ \\
        \multicolumn{10}{c}{~} \\
        Method & \textbf{cp$_9$} & \textbf{cp$_{10}$} & cp$_{13}$ & \textbf{cp$_{14}$} & cp$_{15}$ & \textbf{cp$_{16}$} & cp$_{17}$ & \textbf{cp$_{18}$} & \textbf{cp$_{19}$} \\ \hline \hline
        gemini & 0.16 & 0.11 & 0.00 & 0.39 & 0.02 & 0.16 & 0.08 & 0.38 & 0.11 \\
        gpt & 0.30 & 0.09 & 0.00 & 0.17 & 0.02 & 0.16 & 0.14 & 0.28 & 0.11 \\
        mistral & 0.28 & 0.16 & 0.00 & 0.31 & 0.14 & 0.22 & 0.06 & 0.50 & 0.13 \\ \hline
        \rowcolor{green!25}
        average & \textbf{0.24} & \textbf{0.12} & 0.00 & \textbf{0.29} & 0.06 & \textbf{0.18} & 0.09 & \textbf{0.39} & \textbf{0.11} \\ \hline
        \rowcolor{green!25}
        \makecell{predominant\\pattern} & cp$_{10}$ & \makecell{cp$_{13}$,\\cp$_{19}$} & - & cp$_1$ & - & cp$_9$ & - & cp$_{17}$ &  \makecell{cp$_{9}$,\\cp$_{13}$}  \\
    \end{tabular}
    \label{tab:ttff}
\end{table}

For the patterns \textbf{cp$_3$, cp$_{8.1}$, cp$_{8.2}$, cp$_9$, cp$_{10}$, cp$_{16}$, and cp$_{19}$} we perform another round of evaluation (see Tab.~\ref{tab:last}).

{
     \scriptsize
    \begin{xltabular}{155pt}{>{\scriptsize}c|>{\scriptsize}c|>{\scriptsize}c|>{\scriptsize}c|>{\scriptsize}X}
    \caption{ Critical Inconsistency: Detailed Analysis}
    \label{tab:last} \\

    Pattern & \makecell{Misclas-\\sified As} & \makecell{Inter-\\changeable?} & Cause & Example and Explanation \\ \hline
    \endfirsthead

    \multicolumn{5}{c}%
    {\tablename\ \thetable{}: Critical Inconsistency: Detailed Analysis} \\
    Pattern & \makecell{Misclas-\\sified As} & \makecell{Inter-\\changeable?} & Cause & Example and Explanation \\ \hline
    \endhead
                cp$_3$ & cp$_{10}$ & No & LLM & \textit{``Move task C to be done in both cases if status after task a is true or false, it should still be done before task D.''} \\
        ~ & ~ & ~ & ~ & The LLM associates "move" with something more complex like embedding in a conditional branch due to the presence of conditional references in the wording. \\ \hline
        cp$_{8.1}$ & cp$_{10}$ & No & User & \textit{``The task D could only be conducted if the last condition is false.''} \\ \hline
        cp$_{8.1}$ & cp$_{8.2}$ & No & LLM  & \textit{``Before Task D, add a conditional Split. If true then end, if false then Task D and loop back to the conditional Split.''} \\
        ~ & ~ & ~ & ~ & In most cases the user intent is to conditionally enter a loop where particular task is executed only if a specific condition is met (clearly describing a pre-condition loop structure). The misclassification likely results from the complexity of the redesign request, where the looping behaviour is described in multiple partial sentences, making it difficult to accurately identify the position of the condition evaluation. \\ \hline
        cp$_{8.2}$ & cp$_{10}$ & No & User  & \textit{``Do task D only when condition is true.''} \\ \hline
        cp$_9$ & cp$_{10}$ & No & User  & ``Create task D and create a new branch after task A, so that task D can be executed instead of task B and C.'' \\ \hline
        cp$_{10}$ & cp$_{13}$ & No & LLM , User  & \textit{``If a = 1, the XOR fragment of task B and C should be skipped.''} \\
        ~ & ~ & ~ & ~ & The user did not clearly express the intent to insert a new conditional branch. The LLM interpreted the prompt as a condition update rather than a new conditional structure, due to the absence of explicit wording and lack of LLM's familiarity with the process context.\\ \hline
        cp$_{10}$ & cp$_{19}$ & No &  LLM, User  & \textit{``Please add XOR in front and after the decision option of false and true.''} \\
        ~ & ~ & ~ & ~ & The distinction between "adding" and "modifying" decision logic likely needs to be better defined in the pattern descriptions or clarified through examples. \\ \hline
        cp$_{16}$ & cp$_9$ & No & User  & \textit{``In the new model, the system executes tasks B and E simultaneously.''} \\ \hline
        cp$_{19}$ & cp$_9$  & Partially &  LLM, User  & \textit{``Replace OR with AND", "Instead of XOR condition, make the tasks be performed in parallel (AND condition).''} \\
        ~ & ~ & ~ & ~ & The misclassification results as changing an XOR to an AND gateway can be interpreted as either modifying decision logic or enabling parallel execution. \\ \hline
        cp$_{19}$ & cp$_{13}$  & No & User  & \textit{ ``Make no more conditions for the task B and C.''} \\
        ~ & ~ & ~ & ~ & The misclassification is due to the user pattern misunderstanding, as removing conditions does not automatically imply the need for a parallel gateway but rather a modification in decision logic. \\
    \end{xltabular}
}

\textbf{Summary:}
As can be seen above, there are multiple reasons why conversational model redesign failed. The reasons for that are not only related to the agent and its pattern understanding, but also to the users themselves, their understanding of the patterns, and their application in various scenarios.

Despite the fact that in most cases, either the user or the LLM is at fault when a pattern fails, in some cases it is not straightforward to define whose fault it is. In such cases, we are dealing with ambiguity in the pattern itself, and clarification of the pattern is necessary.

\textbf{Pattern ambiguity} means that the wording can be interpreted as multiple patterns, as the definition of the pattern provided to the LLM is not completely clear and can be interpreted in multiple ways. As a result, the identified pattern may differ from the expected one but cannot be considered a truly wrong classification, since this information was not provided to the LLM from the beginning. In some cases, the change can be realised through multiple patterns, but this is not a consistent setup; rather, it is a coincidence based on the current example or settings.

Based on the most frequent causes for underperformance, all cases discussed above can be classified into the following \rnk{four categories, i.e., \textsl{i) No failures}, \textsl{ii) User}, \textsl{iii) LLM}, and \textsl{iv) Pattern Ambiguity}. This grouping provides a comprehensive foundation for further improvement of agent implementation along with the pattern taxonomy.}

{\nkMulti
The \textsl{i) No failures} group includes patterns with correct behaviour (see Tab.~\ref{tab:matching-existing}) and provides a reference point for evaluating agent performance and defining patterns that perform correctly, requiring less attention and effort for further investigation.

The \textsl{ii) User} group includes patterns that were not identified or where meaning was not derived (see Tabs.~\ref{tab:mapping-false} and \ref{tab:no-meaning}), since due to the analysis performed earlier, the failures in these phases arise mostly from poor user input. \newnk{If the input is unclear or ambiguous, the intended meaning may not be correctly inferred. }

For \textsl{iii) LLM} failures, we consider incorrect pattern implementation (see Tab.~\ref{tab:tttf}), where the LLM fails to understand the pattern and, as a consequence, cannot apply it correctly, even though the other steps were correct.
This category can be used to estimate an LLM’s non-deterministic behaviour across different runs to understand whether the LLM truly reasons and performs the task assigned to it, or just randomly generates output based on provided wordings.

The remaining two groups, (i.e., incorrect pattern application or identification as well as critical inconsistency, see Tabs.~\ref{tab:mismatching} and \ref{tab:ttff}), are classified as \textsl{iv) Pattern Ambiguity}. In these cases, it is particularly complicated to identify who or what is the reason for the failure: solely to the user or the LLM.
Patterns falling in this group need improved definition and clear guidelines for the application.
}

\begin{table}[!ht]
    \centering
    \scriptsize
    \caption{\newnk{Values represent the} Average Distribution of Reasons for Change Pattern Failures}
    \begin{tabular}{l|c|c|c|c|c|c|c|c|c}
        Reason & cp$_1$ & cp$_2$ & cp$_3$ & cp$_4$ & cp$_5$ & cp$_6$ & cp$_7$ & cp$_{8.1}$ & cp$_{8.2}$ \\ \hline \hline
        \makecell{Pattern\\Ambiguity} & 0.17 & 0.38 & 0.18 & 0.08 & 0.05 & 0.29 & \textbf{0.51} & \textbf{0.57} & 0.25 \\
        User & 0.07 & 0.08 & 0.13 & 0.28 & 0.21 & 0.18 & 0.21 & 0.16 & 0.18 \\
        LLM & 0.04 & 0.08 & \textbf{0.43} & 0.05 & 0.03 & \textbf{0.45} & 0.24 & 0.24 & \textbf{0.55} \\
        No Failure & \textbf{0.72} & \textbf{0.46} & 0.26 & \textbf{0.59} & \textbf{0.71} & 0.08 & 0.05 & 0.02 & 0.02 \\
        \multicolumn{10}{c}{~} \\
        Reason & cp$_9$ & cp$_{10}$ & cp$_{13}$ & cp$_{14}$ & cp$_{15}$ & cp$_{16}$ & cp$_{17}$ & cp$_{18}$ & cp$_{19}$ \\ \hline \hline
        \makecell{Pattern\\Ambiguity} & 0.25 & 0.13 & 0.04 & \textbf{0.71} & 0.11 & 0.21 & 0.17 & \textbf{0.61} & 0.20 \\
        User & 0.27 & 0.16 & 0.07 & 0.18 & 0.25 & 0.26 & 0.16 & 0.27 & 0.28 \\
        LLM & \textbf{0.39} & \textbf{0.70} & 0.00 & 0.10 & 0.07 & 0.23 & 0.01 & 0.08 & 0.06 \\
        No Failure & 0.09 & 0.02 & \textbf{0.89} & 0.01 & \textbf{0.57} & \textbf{0.30} & \textbf{0.67} & 0.04 & \textbf{0.46} \\
    \end{tabular}
    \label{tab:final}
\end{table}

As shown in the evaluation, half of the patterns (9 out of 18) performed correctly in the majority of cases (see Tab.~\ref{tab:final}). Four of these patterns were proposed by us and did not exist previously. Based on their high rate of correct outcomes, we can conclude that these patterns are valid candidates to be considered as change patterns for conversational model redesign.

The remaining patterns failed due to issues related to the user, the LLM, or pattern ambiguity (see Tab.~\ref{tab:final}). \nk{However, we cannot immediately suggest the exclusion of these patterns from the CP set based on some thresholds. Each pattern requires further examination to determine whether it still appears to be necessary and, depending on the reason for failure, may need to be clarified either on the agent or the user side.}

\rnk{4 out of 18 patterns fail due to pattern ambiguity. In most cases, either the user provides wrong wording because the pattern itself was not clear to him/her, or an LLM misinterprets both the initial meaning of the pattern and its connection with the wording provided by the user. }

Here, two patterns are particularly interesting, since in some cases, they were able to return process model equal to the ground truth one. In the case of cp$_{14}$, considering the analysis performed earlier (see Tab.~\ref{tab:alternative}), we can see that the misclassification appears to be caused by the user wording, which referred to cp$_1$ instead of cp$_{14}$. Since the suggested alternative, cp$_1$, leads to the same outcome as the expected pattern cp$_{14}$, cp$_1$ can be considered a valid substitute. A similar situation applies to cp$_{18}$, which was effectively realised through cp$_{17}$. In the case of cp$_{18}$, the same result can be indeed achieved by applying cp$_{17}$. Thus, these patterns are completely interchangeable.

However, in the case of pattern cp$_{14}$, there is a substantial difference between copying and inserting a new process fragment, even though at first glance these patterns may appear to produce the same result. When copying, the process fragment remains unchanged, preserving all of its original properties. In contrast, inserting creates a new process fragment from scratch, without inheriting any properties or characteristics from the original.

\rnk{The remaining 5 patterns failed mostly due to the LLM (i.e., cp$_3$, cp$_6$, cp$_{8.2}$, cp$_9$, and cp$_{10}$), indicating that the guidelines we provided to the LLM were not clear. Task clarification, demonstration of examples (e.g., few-shot prompting strategy), etc., along with explanatory analysis of why exactly the LLM failed, are necessary.}

\rnk{Interestingly, none of the patterns failed due to only user misinterpretation. It is always a combination of multiple patterns. In most cases, users provide meaningful input that leads to a valid output, even though it might be different from the expected one. That indicates that as soon as a user understands the pattern, they can clearly express and communicate it to the LLM. }

Comparing the performance across LLMs, we achieve better results and observe more cases where the agent demonstrates correct behaviour using gpt model. \newnk{Gemini shows the worst performance. However, in comparison to mistral, both} gemini and gpt adhere more closely to the instructions and return output that is consistent with the provided guidelines. Mistral, on the other hand, tends to return output (e.g., multiple identified patterns, comments, and explanations) that was not requested, making evaluation more difficult.

However, despite the fact that in some cases gemini performs worse compared to mistral and gpt, the average case distribution in the evaluation remains similar across all three LLMs. This means that, by applying the selected procedure and prompts, we are still able to obtain consistent results, the comprehension capabilities of the LLMs are similar, and our findings can be considered valid.

{\nkMulti
\subsection{Baseline vs. CPMR }

In Table \ref{tab:compare}, we report the average results for the baseline approach, \newnk{along with the average results for the CPMR approach}, including all cases in which the generated models match the ground-truth models (AAO == EAO), even if other intermediate steps fail (e.g., correct behaviour but incorrect pattern identification or application). This allows us to assess whether the filtering mechanisms in the CPMR approach are adequate, since the baseline approach contains no such mechanisms and considers only the similarity between AAO and EAO.

\begin{table}[!ht]
    \centering
    \scriptsize
    \caption{Comparison of Baseline and CPMR Approaches: \newnk{Values represent the} percentage of cases in which EAO equals AAO}
    \begin{tabular}{l|c|c|c|c|c|c|c|c|c}
        Approach & \textbf{cp$_1$} & \textbf{cp$_2$} & cp$_3$ & \textbf{cp$_4$} & \textbf{cp$_5$} & cp$_6$ & cp$_7$ & cp$_{8.1}$ & cp$_{8.2}$ \\ \hline \hline
        Baseline & 0.90 & 0.79 & 0.40 & 0.93 & 0.90 & 0.40 & 0.29 & 0.06 & 0.09 \\
        CPMR & 0.81 & 0.69 & 0.28 & 0.65 & 0.72 & 0.24 & 0.20 & 0.03 & 0.04 \\
        \multicolumn{10}{c}{~} \\
        Approach & cp9 & cp10 & \textbf{cp13} & \textbf{cp14} & \textbf{cp15} & \textbf{cp16} & \textbf{cp17} & cp18 & \textbf{cp19} \\ \hline  \hline
        Baseline & 0.17 & 0.07 & 0.98 & 0.58 & 0.88 & 0.51 & 0.79 & 0.44 & 0.76 \\
        CPMR & 0.09 & 0.03 & 0.93 & 0.42 & 0.63 & 0.33 & 0.74 & 0.27 & 0.55 \\
    \end{tabular}
    \label{tab:compare}
\end{table}

By comparing results from the baseline and CPMR approach, we can see that the performance results \newnk{exhibit the same trend, i.e., patterns that succeeded in the baseline also succeeded for the CPMR approach, although the actual performance of the CPMR approach is lower than the results achieved during the baseline.}

\newnk{The difference in performance between the two approaches is $12$\% on average, with most of the patterns having between $3$--$18$\% of difference. However, for patterns cp$4$, cp${15}$, and cp$_{19}$ the difference exceeds $20$\%. The low performance of cp$_4$ can be explained, since users often utilise the ``rename'' operation instead of ``replace'', where ``rename'' is not a pattern. For the other two patterns, low performance can be explained by the high percentage of failures due to the user (i.e., unidentified patterns or underived meaning, see Tab.~\ref{tab:final}).}

\newnk{We also consider how the two approaches are aligned with each other by calculating, for each pattern, the cases in which the results are the same (i.e., EAO == AAO and EAO != AAO) and the cases in which they are not. Across all patterns, the average agreement rate reaches $82$\% (see Tab.~\ref{tab:agree}), while disagreements remain limited ($18$\% on average), demonstrating strong behavioural similarity between the two approaches. It is also important to mention that we expect a certain level of disagreement, since the baseline approach performs better than the CPMR approach.}

\begin{table}[!ht]
    \centering
        \caption{Alignment of Baseline and CPMR Approaches: \newnk{Values represent the} percentage of cases in which AAOs produced by both approaches are identical}
    \begin{tabular}{l||ccc||c}
        ~ & gemini & gpt & mistral & average \\ \hline
        Agreement & 0.78 & 0.86 & 0.81 & 0.82 \\
        Disagreement & 0.22 & 0.14 & 0.19 & 0.18 \\
    \end{tabular}
    \label{tab:agree}
\end{table}

Overall, similar results with the 12\% difference on average between the two approaches \newnk{and their high alignment with each other} indicate that the filtering mechanisms work successfully. However, due to the limited number of patterns that perform correctly, we can see that filtering mechanisms alone are not enough.
The baseline performs on average better than CPMR, since incorrect input is still processed and there is a chance to get a randomly generated (but correct) model back, since the settings and models we are relying on are rather simplistic. The CPMR approach provides more control, but sacrifices performance.}

\newnk{In general, we observe that even though CPMR involves multiple steps, its complexity does not compromise the performance. Its results remain comparable to the baseline, handling each pattern similarly.
For simple process models (i.e., models with low structural complexity) and straightforward patterns (i.e., patterns with clear and unambiguous modification logic), the baseline approach seems to be sufficient, enabling fast modifications. The CPMR approach provides an extensible framework allowing incorporation of user input verification to prevent incomplete or ambiguous changes (particularly in cases where partially specified modifications could lead to errors or constraint violations) and supporting feedback generation to guide users to enrich their input. Finally, the CPMR approach offers a higher level of control, providing assurance that the agent (and associated with it the LLM) applies changes exactly as intended rather than succeeding accidentally or by chance from time to time.}

\subsection{Discussion}

{\nkMulti
One insight from the evaluation is that independent of the approach, there are patterns that perform correctly and patterns that systematically fail. The reasons for these failures are mostly due to either the agent or to pattern ambiguity.

For the patterns cp$3$, cp$6$, cp${8.2}$, cp$9$, and cp$_{10}$ that failed because of the LLM, we need to improve the utilised prompts and consider changing the prompting strategy by adding examples, clarifying or even formalizing the pattern descriptions, or modifying the architecture of the current pipeline. This could involve combining LLM capabilities with traditional deterministic approaches to improve model redesign performance. For instance, the LLM could be used to identify the pattern and extract relevant parameters from the wording, while the pattern application itself could be carried out using deterministic methods.

The high presence of pattern ambiguity in some cases indicates that pattern clarification is required not only for the agent, but also for the user, to minimise misunderstandings and incorrect pattern application from their side.
\newnk{Few-shot prompts with emphasis on pattern ambiguity can help the LLM better understand task-specific nuances, reduce ambiguity in pattern recognition, and improve consistency across multiple patterns.}

\srmrev{Patterns rarely failed because of wrong user input. This suggests that as soon as users understand a pattern, they can apply it correctly.
These and previous findings highlight the necessity of a recommender system implemented by the agent to support the user in wording clarification, especially in the beginning of the redesign interaction, to achieve better results.}

This is also consistent with the results obtained from the comparison of the baseline against the CPMR approach. The baseline approach tends to have better results on average, but also provides less control and fewer opportunities to determine the reasons for failure. Thus, in the future, to improve the results, it is necessary not only to improve agent implementation or communicate patterns to the users more easily, but also to reconsider the existing definition of patterns (i.e., determine whether some old patterns are unnecessary, or whether some new patterns not considered in this work are necessary). Moreover, we recommend to apply a \textbf{hybrid approach}, where the agent does not simply stop after facing inaccurate user input or pattern ambiguity, but can generate follow-up questions or recommendations based on its change pattern knowledge.
}

\newnk{The evaluation was conducted in a domain-agnostic setting using contextless examples. Introducing domain-specific terminology may affect the results in several ways, e.g., the performance and the number of patterns that perform successfully, and require additional considerations to achieve improvements. For example, improving performance would require not only incorporating pattern-related information into the prompt, but also introducing domain-specific knowledge to help the agent better understand the underlying context.}

\rnk{Given multiple wordings for one particular pattern, with different styles and from different users ($64$ in total for each pattern), and the observation that these distinct wordings for the same change pattern consistently lead to correct (and for some patterns also incorrect) outputs across multiple LLMs using different approaches (baseline and CDP), we can conclude that the agent and the LLMs in the background are robust and mostly insensitive to user wording variations. This means that as soon as the pattern is clearly expressed by the user and is understandable for the LLM, the LLM can understand it regardless of linguistic differences.}

Interestingly, when considering user wording in general regardless of the specific pattern, there are two common user tendencies that occurred.
First, when the changes required for the models were too complex or lengthy to describe, users tended to create a new model from scratch (e.g., ``Delete everything and create a new process with task A, then task B''). Second, in some cases, users employed the concept of reverting changes, referring back to a previous model state (e.g., ``Reverse the action from the previous question/step''). These two behaviours could also be considered as patterns, not in terms of change patterns for model refinement, but rather as patterns in user behaviour that have to be supported by the agent implementation.

\rnk{Comparing traditional modelling tools with the proposed approach, we can state that traditional approaches usually check only whether, after the changes performed by the user, the model is still syntactically correct and adheres to the modelling guidelines. However, the semantic meaning of changes is not considered (since it is also not possible, as the expected user changes exist only in the user's mind). In CPMR, we are able to evaluate whether the expectations of the user are truly adhered into the changes obtained at the end. This gives a new perspective and insights on how the changes are executed, how users actually interpret and apply them, where misunderstandings or systematic failures occur, and how the set of available patterns could be improved.}

\subsection{Threats to validity}

This study presents several limitations that may affect the generalizability and validity of the findings.

The process models used in the evaluation are small and simple, consisting of generic task labels lacking domain-specific semantics (e.g., Task A, Task B, etc.). While this design choice ensures clarity and consistency across all participants and LLMs, it limits the applicability of the results to more complex scenarios as our examples may not fully represent the challenges users face when working with larger or more complex process models.

Participants were not engaged in an interactive setting with the conversational agent. Instead, they were asked to provide wording based on a static input-output model pair. Consequently, user behaviour in this study may differ from that observed in real deployment contexts.

Users were required to express their intent using a single prompt per model transformation. This constraint does not reflect natural user-agent interaction patterns, where clarification and follow-up are common.

\rnk{In the CPMR approach, one of the underlying assumptions is that users address tasks by their labels with the expectation that these labels are unique within each process model. We do not know how the approach behaves when duplicate labels are present. In such cases, it might be reasonable to address tasks by their IDs. However, this would require the IDs to be visualised, which is generally not common for the BPMN standard.}

In addition, the output models used in the evaluation were constructed based on a predefined set of change patterns. While this approach allowed for better evaluation and comparison, it may not fully reflect the variety of real-world redesign requests. In real-world scenarios, users may express additional or more complex transformation goals that were not anticipated or covered by the selected patterns.

LLM behaviour can vary based on version, update timing, and underlying training data. Future replications may observe different results due to changes in LLM behaviour.

\section{Related Work}
\label{sec:rewo}

Business process modelling requires the accurate representation of intricate workflows, decision points, and interactions between multiple stakeholders within an organization. This complexity is further compounded by the need for effective communication between domain experts, who possess the contextual knowledge, and process modellers, who translate that knowledge into formal models, as well as the need to update and redesign the model in the future.

Several studies address the communication gap between domain experts and process modellers, e.g., \cite{labeling,seven,guide,Approximating,SBVR}. With recent advancements in natural language processing and generative AI, which are transforming transforming classical BPM systems into AI-augmented BPM systems~\cite{next-gen}, the use of natural language and chatbots can be a realistic option for overcoming the communication gap.

These systems become conversationally actionable, meaning they can proactively communicate with human agents about process-related actions, goals, and intentions using natural language~\cite{manifesto}. This interaction can be enhanced via the integration of intelligent chatbot functions for improved communication within the BPM framework, promoting collaboration~\cite{next-gen}. The systems can lead conversations in a multi-turn nature, considering context and incorporating utterances from previous turns to achieve a higher degree of user engagement~\cite{conversys}.
Currently, as mentioned in~\cite{LPM,PromptEng,ProcessGPT,VidgofBM23,chit}, there is an increasing interest in the potential benefits for the entire BPM domain arising from employing LLMs, particularly in process model generation.
However, most existing approaches focus solely on single-time interactions, where the user is able to receive a final artifact from the system, but is not able to adjust and redesign it.
So far, the multi-turn conversational capabilities of LLMs for process modelling have received little attention and have not yet been thoroughly explored in the BPM domain.

A process model can be redesigned by rearranging various elements to satisfy predefined business rules and constraints. The primary goal of process redesign is process improvement, which can be categorised into two levels: (a) functional goals, such as ensuring desired or acceptable process behaviour, and (b) non-functional goals, including cost reduction, time optimization, quality of service enhancement, and increased flexibility~\cite{9241415}.

To facilitate process redesign, researchers have introduced process improvement/redesign patterns (PIPs)—generic concepts aimed at enhancing specific aspects of business processes~\cite{improvement-patterns}. Since the early 2000s, numerous studies have focused on process redesign strategies, process redesign patterns and frameworks. A comprehensive overview of these patterns can be found in~\cite{framework}. These works demonstrate that redesign patterns cover multiple aspects of process models, ranging from structural changes and data transformations to quality, compliance, and risk-related modifications.

However, identifying the most suitable redesign pattern for a given scenario remains challenging, as most researchers focus on only a subset of available patterns, driven by specific requirements and constraints. Given that the scope of our paper is to explore structural changes in process models using process redesign patterns, we refer specifically to change patterns. Several studies have examined change patterns in the context of structural modifications and model variability~\cite{vari,kim,framework,redesign}.

In our work, we adopt change patterns as a foundation, specifically referring to the adaptation patterns introduced by Weber et al., as these patterns are widely recognised and well-established in literature~\cite{pattern-survey}. Change patterns support users in performing structural modifications on process models by encapsulating multiple low-level actions on individual model element (e.g., add node, delete node, add edge, remove edge, move edge) into a single, semantically meaningful transformation, simplifying process modifications for users. These high-level change operations provide a higher level of abstraction, enabling complex transformations that maintain process integrity~\cite{weber}.

Currently, despite recent interest in AI-augmented BPM systems and the conversational capabilities of LLMs, \rnk{only few approaches attempt to incorporate user feedback into process models~\cite{promAI,DBLP:conf/bpm/KopkeS24a}, yet} the potential of generative AI to support multi-turn, user-guided process model redesign using change patterns remains under-explored.

\section{Conclusions}
\label{sec:conclusion}
In this work, we explore whether an LLM-based agent can effectively support domain experts during the redesign of process models in continuous interaction via a conversational user interface, aiming to overcome the communication gap between domain experts and process modelers. The continuous interaction is based on redesign tasks of the models. To this end, we \srm{propose a conversational process model redesign approach that incorporates and adapts} existing change patterns for business process models \srm{into a conversational context}. The approach is systematically evaluated against process models that are redesigned by the user and manually using the change patterns.

\nk{We not only evaluate existing change patterns, but we also propose the potential extension of already existing ones, which are necessary due to the conversational nature of the user interface of the LLM-based agent. In addition, we introduce the multifaceted evaluation concept, which allows us to grasp not only the fact of the failure of the model redesign, but also the reason why the failure happened, and show, in practice, how to utilise this methodology through an example with three different LLMs.}

\nk{Model redesign evaluation is a complex task that depends on multiple factors: the user, the LLM, and the nature of the pattern. These include aspects such as how the redesign task is described, whether the description is complete, whether the user applies the correct pattern for the intended change, whether the LLM explicitly understands the provided patterns, which patterns appear similar even if they are not, and the complexity of the change itself.}

\nk{On average, 9\% of cases failed at the stage of identification and 2\% at the stage of meaning derivation due to poor wording, highlighting the need for mechanisms that support users in minimizing ambiguity, improving clarity, and selecting the appropriate pattern for their request.}

\nk{On the other hand, the high level of failures due to the agent during pattern application, especially in more complex cases, indicated the necessity of improved methods of pattern application (i.e., hybrid approaches leveraging the strengths of both, LLMs and traditional approaches).}

\rnk{Most of the patterns fail due to pattern ambiguity, indicating that some patterns are complicated for both, users and LLMs. This shows that, on one hand, it is necessary to improve the clarity of these patterns, and on the other hand, the system should be able to generate follow-up questions to improve the quality of the obtained user input.}

\rnk{Overall, based on the results, a \textsl{hybrid} CPMR approach seems to be an opportunity to bridge the advantages of both approaches (e.g., higher performance and control) by first identifying intended change patterns, followed by the direct application of those patterns that work correctly, and for others, follow-up questions that are to be derived to improve user input.}

\rnk{Future research could focus on four  directions. The first may explore more complex datasets to address the increasing complexity of real-world scenarios.
The second one might focus on the analysis and refinement of the change pattern taxonomy, defining change pattern overlap and redundancy, or pattern granularity, and evaluating whether this refinement influences performance. }
The third direction could focus on the implementation of change patterns utilizing formalization, i.e., deriving meaning not in a free natural language form but rather in a formal way, to perform its implementation via traditional approaches. This will minimise deviations between expected and actual agent output, helping to focus on conversational and human-agent interaction aspects. The last one could emphasise evaluating and integrating knowledge about user behaviour to improve the quality of human-chatbot communication, better meeting the needs of domain experts. Additionally, observing this communication as a learning process for domain experts may help develop their modelling skills and foster \textit{process thinking} through active engagement in process model creation.

\bibliographystyle{ws-ijcis}
\bibliography{bib.bib}

\end{document}